\pdfoutput=1

\documentclass[11pt]{article}

\usepackage[]{acl}
\usepackage{gb4e}
\noautomath
\usepackage{paralist}
\usepackage{mathtools}
\usepackage{times}
\usepackage{latexsym}
\usepackage{amsmath}
\usepackage[capitalize]{cleveref}
\usepackage{amssymb}
\usepackage{multirow,multicol}
\usepackage{graphicx}

\usepackage{array}
\usepackage{bm}
\usepackage{soul, color}
\usepackage{booktabs}
\usepackage[colorinlistoftodos]{todonotes}
\usepackage{amsthm}
\usepackage{caption}
\usepackage{subcaption}
\usepackage{tipa}
\usepackage{stackengine}
\usepackage{rotating}
\usepackage{tabularx}
\usepackage{dialogue}
\usepackage{enumerate}
\usepackage{enumitem}
\usepackage{txfonts}
\usepackage{colortbl}
\usepackage{makecell}
\usepackage{yfonts}
\usepackage{float}

\newcommand\code[1]{\texttt{#1}}
\crefformat{section}{\S#2#1#3} 
\crefformat{subsection}{\S#2#1#3}
\crefformat{subsubsection}{\S#2#1#3}
\newcommand{\declarelogo}[0]{\includegraphics[height=.02\textwidth]{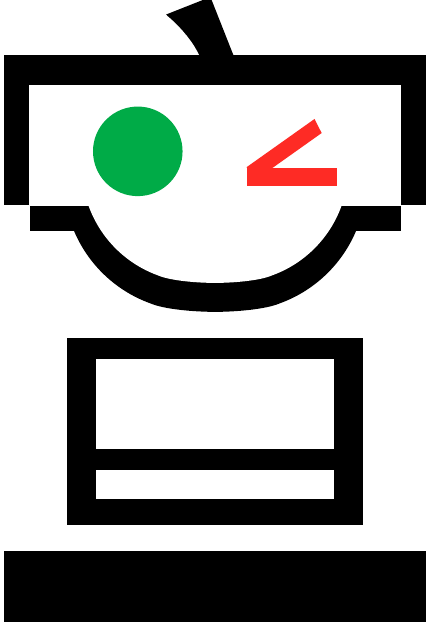}}
\newcommand{\umichlogo}[0]{\includegraphics[height=.012\textwidth]{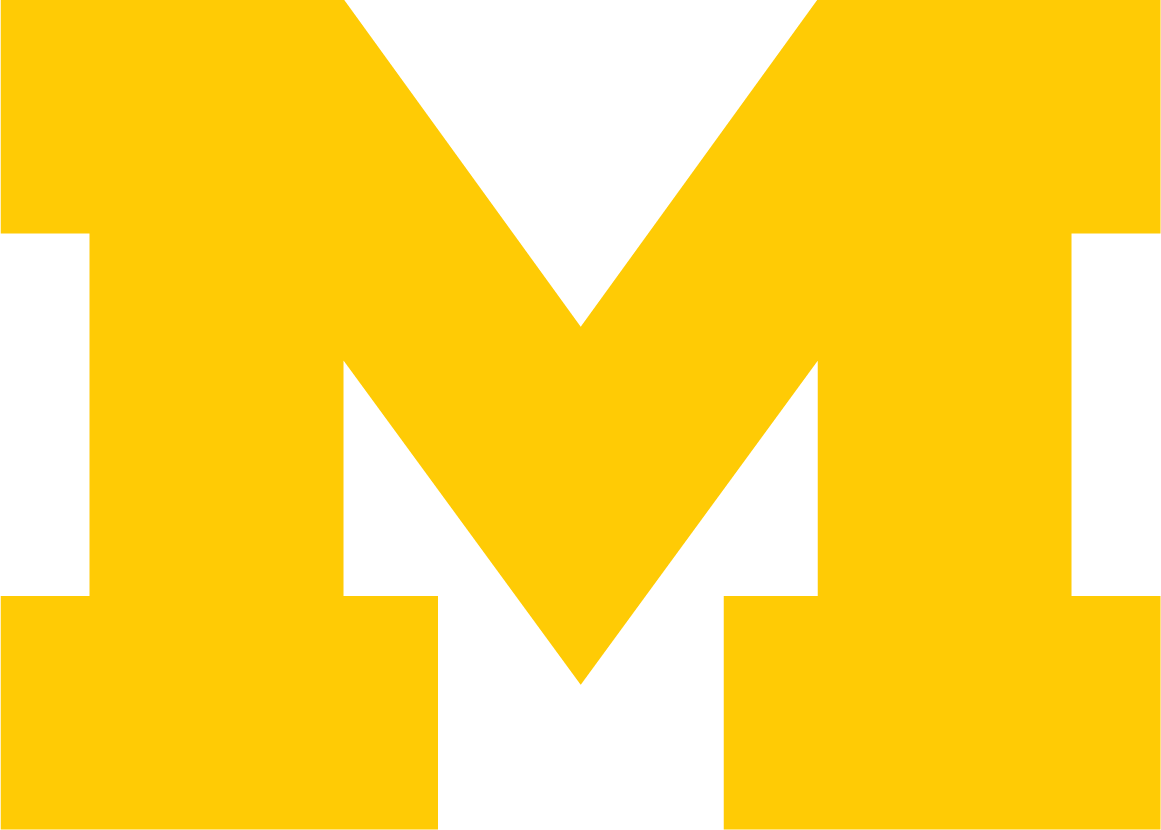}}

\newcommand\datatwofont[1]{{\usefont{T1}{cinzeldecorative}{m}{n}#1}}
\newcommand\datatwofonttitle[1]{{{\usefont{T1}{cinzeldecorativebold}{m}{n}#1}}}

\newcommand{\dataset}{{\datatwofont{CICERO}}}
\newcommand{\newdatasetnamelong}{\datatwofonttitle{C}ontextual\datatwofonttitle{I}zed \datatwofonttitle{C}ommons\datatwofonttitle{E}nse Infe\datatwofonttitle{R}ence in dial\datatwofonttitle{O}gues}
\newcommand{\newdatasetnametitle}{{\datatwofonttitle{CICERO}}}

\newcommand{\rotatetabular}[4]{\multirow{#1}{*}{\rotatebox{90}{\colorbox{#2}{\begin{tabular}{c}\tiny \bf #3\\ \tiny \bf #4\end{tabular}}}}}
\newcommand{\rotatetabularnormal}[4]{\multirow{#1}{*}{\rotatebox{90}{\colorbox{#2}{\begin{tabular}{c}\scriptsize \bf #3\\ \scriptsize \bf #4\end{tabular}}}}}

\usepackage{pifont}

\definecolor{darkpastelgreen}{rgb}{0.01, 0.75, 0.24}
\definecolor{brilliantlavender}{rgb}{0.96, 0.73, 1.0}
\definecolor{Gray1}{rgb}{0.91,0.925, 0.937}
\definecolor{cambridgeblue}{rgb}{0.0, 0.8, 0.6}
\definecolor{Gray2}{rgb}{0.87, 0.886, 0.902}
\definecolor{Gray3}{rgb}{0.808, 0.831, 0.855}
\definecolor{Gray4}{rgb}{0.678,0.71, 0.741}
\definecolor{darkgreen}{rgb}{0.0, 0.5, 0.0}
\definecolor{almond}{rgb}{0.99, 0.87, 0.9}
\definecolor{ghostwhite}{rgb}{0.98, 0.81, 0.69}
\definecolor{Blue1}{rgb}{0.792, 0.941, 0.973}
\definecolor{Blue2}{rgb}{0.678, 0.91, 0.957}
\definecolor{Blue3}{rgb}{0.565, 0.878, 0.937}
\definecolor{Blue4}{rgb}{0.282, 0.749, 0.89}
\definecolor{Yellow1}{rgb}{1, 0.914, 0.306}
\definecolor{Yellow2}{rgb}{1, 0.886, 0.275}
\definecolor{Yellow3}{rgb}{1, 0.855, 0.239}
\definecolor{gold}{rgb}{0.85,.66,0}
\definecolor{aqua}{rgb}{0.80784314, 0.90196078, 0.35294118}
\definecolor{Green0}{rgb}{0.909, 0.992, 0.886}
\definecolor{Green1}{rgb}{0.843, 0.960, 0.839}
\definecolor{Green2}{rgb}{0.635, 0.854, 0.627}

\definecolor{Amber}{rgb}{0.99, 0.76, 0.8}
\definecolor{bluebell}{rgb}{0.64, 0.64, 0.82}

\title{\newdatasetnametitle: A Dataset for Contextualized\\ Commonsense Inference in Dialogues}
\date{}

\author{Deepanway Ghosal$^{\declarelogo}$ \hspace{2mm}
Siqi Shen$^{\umichlogo}$ \hspace{2mm}
Navonil Majumder$^{\declarelogo}$ \hspace{2mm}\\
\textbf{Rada Mihalcea$^{\umichlogo}$ \hspace{2mm}
Soujanya Poria$^{\declarelogo}$}\\
  $^{\declarelogo}$ DeCLaRe Lab, Singapore University of Technology and Design, Singapore\\
  $^{\umichlogo}$ University of Michigan, USA\\
  \texttt{\{deepanway\_ghosal@mymail.,navonil\_majumder@,sporia@\}sutd.edu.sg}\\ \texttt{\{shensq,mihalcea\}@umich.edu}\\
 \vspace{1mm}\code{\textbf{\dataset{} is available at: \url{https://declare-lab.github.io/CICERO}}}
  }
 
\begin{document}
\maketitle

\begin{abstract}

This paper addresses the problem of dialogue reasoning with contextualized commonsense inference. We curate \dataset{}, a dataset of dyadic conversations with five types of utterance-level reasoning-based inferences: cause, subsequent event, prerequisite, motivation, and emotional reaction. The dataset contains 53,105 of such inferences from 5,672 dialogues. We use this dataset to solve relevant generative and discriminative tasks: generation of cause and subsequent event; generation of prerequisite, motivation, and listener's emotional reaction; and selection of plausible alternatives. Our results ascertain the value of such dialogue-centric commonsense knowledge datasets. It is our hope that \dataset{} will open new research avenues into commonsense-based dialogue reasoning.
\end{abstract}

\section{Introduction}
\label{sec:intro}

Conversational content on the internet is quickly growing, and such content holds valuable knowledge about how information exchange takes place among speakers. A key step towards understanding such dialogues is gaining the ability to reason with the information shared in the dialogue. To this end, we curate a dataset of dyadic conversations named \dataset{}
 (\newdatasetnamelong)
, which contains inferences around the utterances in the dialogues. The dataset focuses on five types of reasoning-based inferences for a given utterance in a dialogue: cause, subsequent event, prerequisite, motivation, and emotional reaction.

\begin{figure*}[t]
    \centering
\begin{subfigure}{0.9\textwidth}
    \centering
        \includegraphics[width=\linewidth]{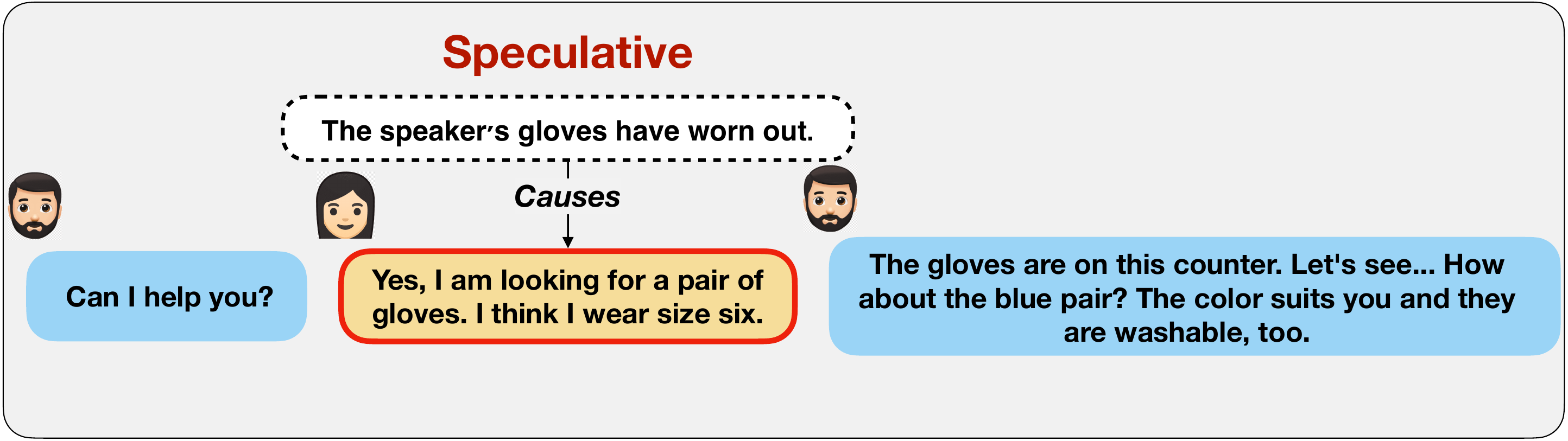}
        \caption{}
        \label{fig:spec}
    \end{subfigure}
    ~
    \begin{subfigure}{0.9\textwidth}
        \centering
        \includegraphics[width=\linewidth]{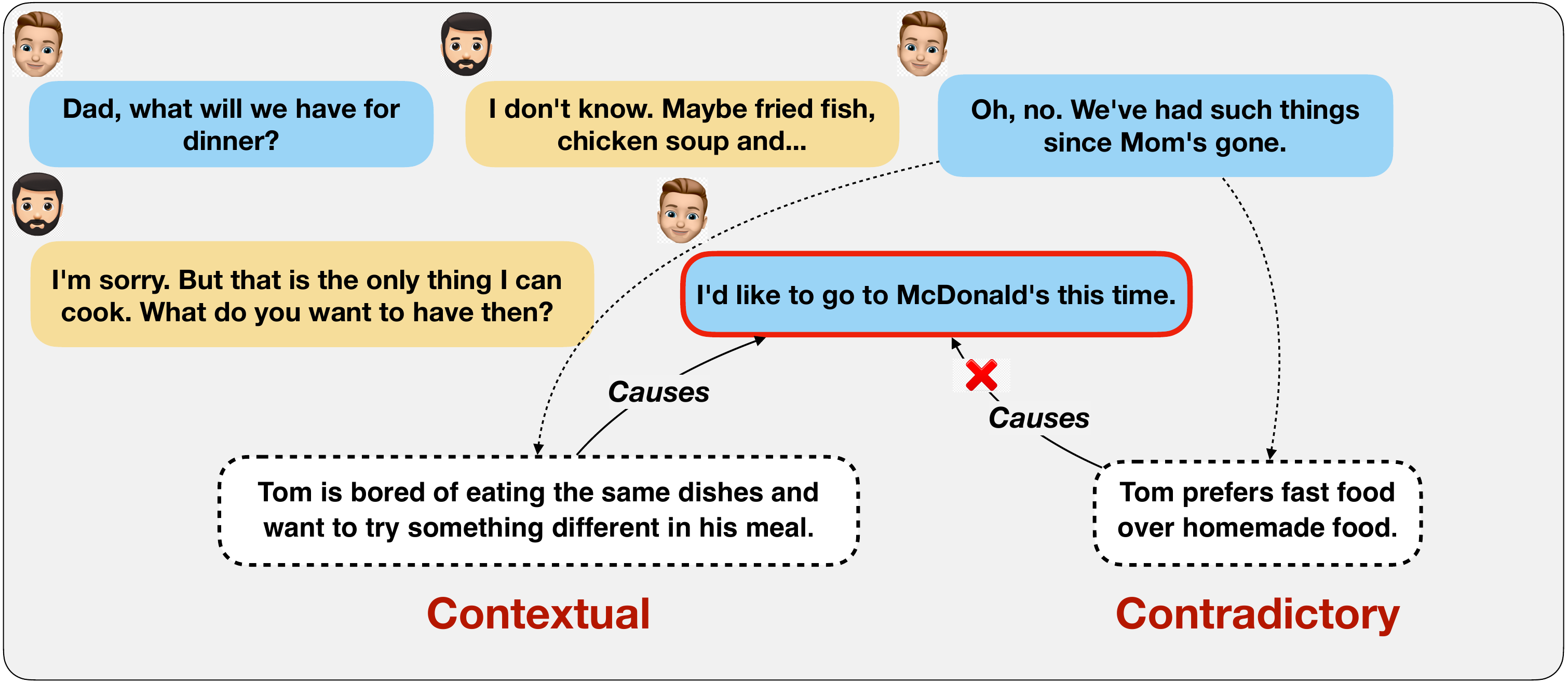}
        \caption{}
        \label{fig:cont}
    \end{subfigure}
    \caption{Illustration of (a) contextualized commonsense speculation and (b) contradictory inferences in dialogues.}
    \label{fig:my_label}
\end{figure*}
Arguably, making such reasoning-based inferences often demands commonsense knowledge, especially when the inference is implicit. \cref{fig:spec} shows such a case where the cause behind the target utterance is not explicit in the context. However, applying the commonsense knowledge \texttt{ worn gloves $\xrightarrow{motivates}$ buy new pair of gloves} allowed the annotator to infer a probable cause of the utterance. On the other hand, commonsense can be crucial in sifting relevant information from the context. \cref{fig:cont} depicts an instance where the cause behind the target utterance is inferred from the context. This inference can be explained by commonsense knowledge (see \cref{fig:intermediate}) such as \texttt{repetitive consumption of the same food $\xrightarrow[]{causes}$ boredom $\xrightarrow[]{dispelled~by}$ changing food $\xrightarrow[]{achieved~by}$ eating at McDonald's}. Thus, it is reasonable to posit that such knowledge could aid to bridge the gap between the input and the target inference.

ATOMIC~\cite{sap2019atomic,hwang2020atomic} is one such dataset for commonsense reasoning-based inference, allowing for a large set of inference types. However, ATOMIC is context-free, as it only provides inferences on short phrases, ignoring the broader context around them. Making an inference on an entire utterance, on the other hand, requires understanding the context around it. As per Grice's maxim~\cite{grice75logic}, in conversations, the interlocutors provide any piece of information as is needed, and no more. Thus, much of the information required to understand an utterance is likely interspersed along the dialogue, and not necessarily localized in the given utterance. For instance, in the example in Figure \ref{fig:cont}, understanding the cause for one of the speakers' desire to go to McDonald's requires the context of the previous utterances. ATOMIC is thus not ideal for commonsense reasoning-based inferences on dialogues, where context is critical for understanding an utterance's implications. We confirm this with our experiments in the subsequent sections (\cref{sec:exp}).

GLUCOSE~\cite{mostafazadeh-etal-2020-glucose} exclusively curates causal inferences –– \emph{cause, enable,} and \emph{result in} -- from monologues. Thus, it is not ideal for making context-consonant inferences on the dialogues. Also, dialogue-specific dimensions like \emph{motivation} and \emph{reaction} are beyond its scope.

On the other hand, CIDER~\cite{ghosal-etal-2021-cider} does provide a dataset for commonsense-based inference on dialogues, but it is limited to inferences explicitly observable in the dialogues. As such, systems based on CIDER cannot effectively speculate around the dialogue for implicit inference.

\dataset{} strives to bring the best of these three datasets by creating a dataset that can enable models to effectively operate on a dialogue by considering the context and speculating when the answer is not apparent.

\section{Construction of \dataset{}}
\label{sec:dataset}

We create \dataset{} -- a large dataset of English dyadic conversations annotated with five types of inferences with the help of human annotators, who are instructed with a carefully crafted set of guidelines.

\subsection{Annotation Instructions}
\label{sec:questions}
\begin{figure}
    \centering
    \includegraphics[width=0.9\linewidth]{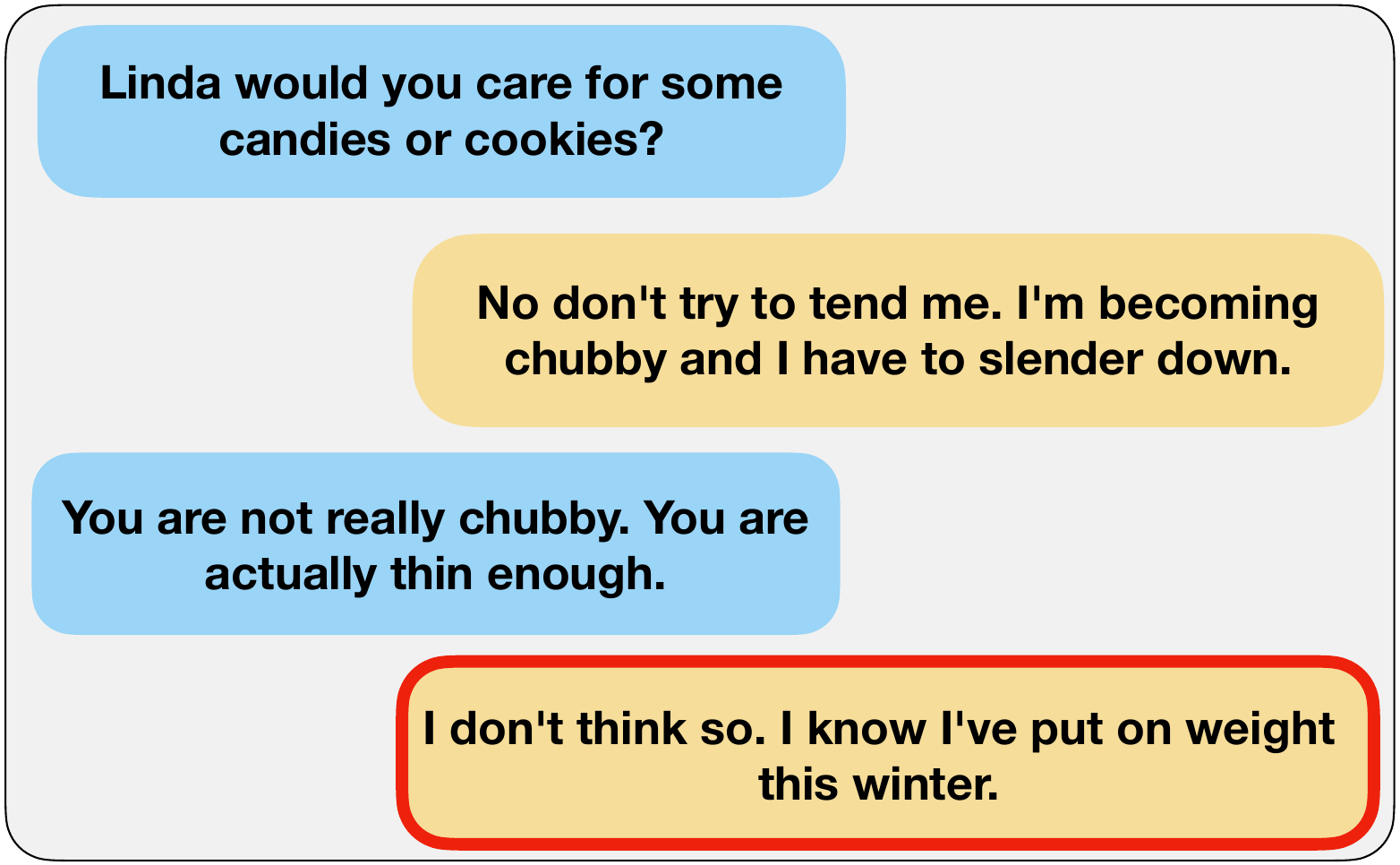}
    \caption{ A \texttt{dialogue-target} pair. The utterances with red border is the target for this dialogue.}
    \label{fig:dialogue-target-1}
\end{figure}

The annotators are given a dialogue and a \texttt{target} utterance, as exemplified in \cref{fig:dialogue-target-1}.
The annotators are then asked to make an inference, posed as a question, about the \texttt{target} utterance. They write a one-sentence answer that is grammatically correct, concise, and consistent with the dialogue. The answer may contain both \emph{overt} and \emph{speculative} scenarios.
An overt scenario is explicitly or implicitly present in the dialogue context. If such contextual scenarios answer the question, the annotators write them as a well-formed sentence. However, in many cases, the dialogue may not hold the answer, neither explicitly nor implicitly. In such cases, the annotators are asked to speculate plausible scenarios around the dialogue, using commonsense and world knowledge, to devise answers that do \texttt{not contradict} the given dialogue context. 

Given the \texttt{dialogue-target} pair in \cref{fig:dialogue-target-1}, at least one of the following five inferences about the \texttt{target} is made by the annotators:

\begin{figure*}[ht]
    \centering
        \includegraphics[width=\linewidth]{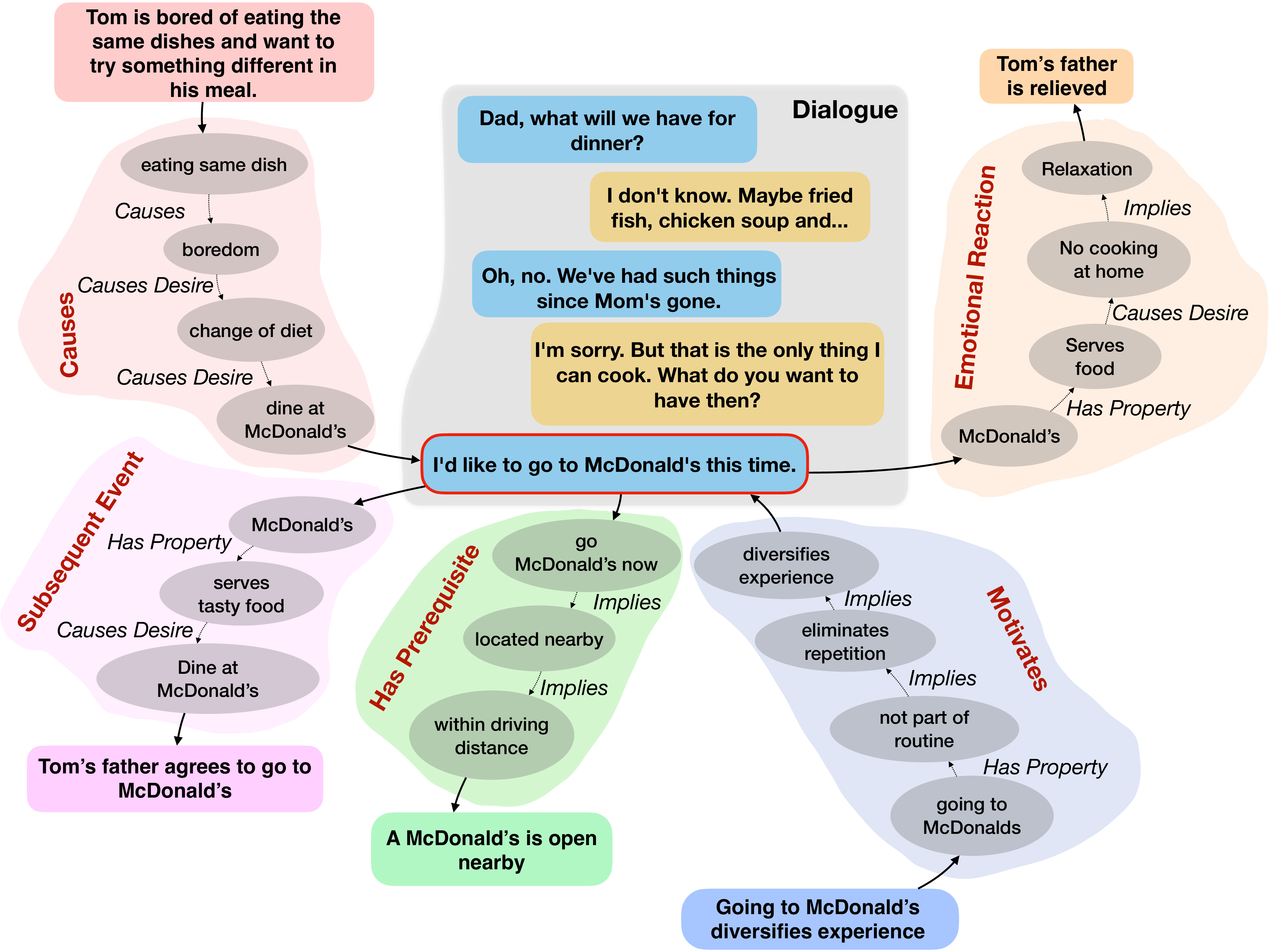}
        \caption{Intermediate commonsense inference steps.}
        \label{fig:intermediate}
\end{figure*}

\paragraph{Q1. } \textbf{What is the event that directly causes (overt) or could cause (speculative) \texttt{Target}?} The annotators consider if any of the events that are or likely to be antecedent to the target can cause the \texttt{target}.

\noindent{\bf Answer:} Linda didn’t exercise regularly during the winter.
\noindent{\bf Remark:} The annotators provided possible, speculative answers as the dialogue itself does not provide any reason for Linda's weight gain.

\paragraph{Q2. } \textbf{What subsequent event happens (overt) or could happen (speculative) following the \texttt{Target}?}
The annotators write about the event that happens or could happen following the \texttt{target}. Additionally, annotators were told that sometimes, such subsequent events of the \texttt{target} are triggered or likely to be triggered by the \texttt{target}.

\noindent{\bf Answer:} Linda starts a diet and tries to lose weight.

\noindent{\bf Remark:} The answer is speculative as the dialogue contains no explicit/implicit subsequent event.

\paragraph{Q3. } \textbf{What is (overt) or could be (speculative) the prerequisite of \texttt{Target}?} Does the \texttt{target} have any direct prerequisite or dependency that has to happen or be fulfilled first? (In most cases, prerequisite is the state/event which has to be satisfied before another event causes \texttt{target}.) The answer is a state/event which enables the happening of the \texttt{target}. In other words, prerequisites are the prior assumptions or background information that the interlocutors agree on about the context.

\noindent{\bf Answer:} Linda was slimmer before the winter.

\noindent{\bf Remark:} Annotators were required to understand the difference between cause and prerequisite clearly before proceeding with the final annotation. Cause of an event X is the event that directly causes X. Prerequisite of an event X is the condition which has to be satisfied in order for X to happen.

\paragraph{Q4.} \textbf{What is an emotion or basic human drive that motivates or could motivate \texttt{Target}?}
Consider the basic human drives, needs (and/or likely emotions) of the speaker of the \texttt{target}. Basic human drives and needs are food, water, clothing, warmth, rest, security, safety, intimate relationships, friends, prestige, feeling of accomplishment, self-fulfillment, creative activities, enjoyment, etc. 
Do any of these human drives/states of mind/emotional feelings motivate the \texttt{target}?

\noindent{\bf Answer:} \code{Not Applicable} for this target. 

\paragraph{Q5.} \textbf{What is the possible emotional reaction of the listener: A (or B)?}
What could be the possible emotional reaction or responses of the listener with respect to the \texttt{target}? The annotators capture the appropriate emotion of the listener using the emotion terms listed in \cref{tab:emotions} verbatim or related words (e.g., anxious, confused, interested, etc). 

\noindent{\bf Answer:} The listener encourages Linda to maintain her diet.

\begin{table*}[t]
\centering
\resizebox{0.8\linewidth}{!}{
	\begin{tabular}{r@{~~}|r@{~~}|r@{~~}|r@{~~}|r@{~~}}
	\toprule
	Admiration & Affection & Afraid & Angry & Annoyed \\
    Anticipating & Anxious & Apprehensive & Ashamed & Awe \\
    Awkwardness & Boredom & Calmness & Caring & Confident \\
    Confusion & Content & Craving & Devastated & Disappointed \\
    Disgusted & Eagerness & Embarrassed & Encouragement & Enthusiasm \\
    Excited & Faithful & Fear & Furious & Grateful \\
    Gratitude & Guilty & Happy & Hopeful & Impressed \\
    Interest & Jealous & Joyful & Lonely & Nostalgic \\
    Prepared & Proud & Relief & Romance & Sad \\
    Satisfaction & Sentimental & Surprised & Terrified & Trusting \\
    \bottomrule
	\end{tabular}
	}
	\caption{Possible emotional reactions of the listener.}
	\label{tab:emotions}
\end{table*}

\paragraph{Additional Guidelines.}
To ensure the quality and diversity of the samples, we also ask the annotators to adhere to the following guidelines:
\begin{itemize}[leftmargin=*]
\setlength\itemsep{-0.25em}
\item Be creative in speculation. Refrain from rephrasing the \textit{target} and writing low-effort trivial answers. It is recommended to skip a question if rephrasing the \textit{target} is the only possible answer.
\item Avoid repeating the same answer for distinct questions on the same \emph{target}.
\item The answer must be consistent with the given dialogue.
\item It is recommended to base the answer on the most important phrase of the \emph{target} should it contain multiple phrases.
\end{itemize}

\subsection{Dialogue Selection for \dataset{}}

\subsubsection{Source Datasets}
To build \dataset{}, we use the dyadic dialogues of the following three datasets:

\noindent \textbf{DailyDialog}~\cite{li2017dailydialog} covers dialogues from wide range of topics --- life, work, relationships, tourism, finance, etc. The constituent utterances are labelled with emotion and dialogue-act.

\noindent \textbf{MuTual}~\cite{mutual} is a multi-turn dialogue reasoning dataset. Given a dialogue history, the objective is to predict the next utterance by considering aspects such as intent, attitude, algebraic, multi-fact, and situation reasoning.

\noindent  \textbf{DREAM}~\cite{sun2019dream} is a multiple-choice reading-comprehension dataset collected from exams of English as a foreign language. The dataset presents significant challenges as many answers are non-extractive and require commonsense knowledge and multi-sentence reasoning.

\subsubsection{Selection Process}
We use the following procedure to select a subset of dialogues from the three datasets:
\begin{enumerate}[leftmargin=*, wide, itemsep=0em, labelwidth=!, labelindent=0pt]
    \item We remove dialogues that are too short or long on either utterance or word level. Dialogues with fewer than five utterances or fewer than six words per utterance on average are removed. Dialogues having more than $15$ utterances or more than $275$ words in total are also removed.
    \item All three source datasets contain dialogues having near identical utterances. 
    We remove these near duplicate dialogues to ensure topical diversity of \dataset{}. We use a sentence embedding model based on fine-tuned RoBERTa~\cite{gao2021simcse} to extract dense feature vectors of the dialogues. We remove the duplicates assuming that a pair of duplicate dialogues have at least $0.87$ cosine similarity.
\end{enumerate}

\subsection{Target Utterance Selection}
\label{sec:target-selection}
Given a dialogue $D$, we select the target utterances as follows:

\vspace{-\topsep}
\begin{itemize}[leftmargin=*, wide, itemsep=0pt, labelwidth=!, labelindent=0pt]
\setlength\itemsep{-0.25em}
    \item We first determine the number of target utterances in $D$: if $D$ has 1--6 utterances, then we select 2 or 3 targets; if it has 7--12 utterances then we select 3--5 targets; otherwise, we select 4--7 targets if it has more than 12 utterances.
    \item We divide $D$ into 2--3 segments having roughly equal number of consecutive utterances. 
    We choose roughly an equal number of the top-ranking utterances from each segment. We call this set of utterances $x_1$. The ranking is performed using a sentence ranking algorithm~\cite{erkan2004lexrank,mihalcea2004textrank} with sentence-BERT embeddings~\cite{reimers2019sentence}.
    \item We also select the longest utterances in $D$ and the utterances that contain phrases such as \textit{I'm, I'd, I've, I'll} or their expansions. We call this set of utterances $x_2$. The sets $x_1$ and $x_2$ may not be disjoint.
    \item Set $x_3$ consisting of the final utterance of $D$. 
\end{itemize}
\vspace{-\topsep}

We choose the inference type for the target utterances from the sets $x_{1,2,3}$ as follows:
\vspace{-\topsep}
\begin{itemize}[leftmargin=*, wide, itemsep=0em, labelwidth=!, labelindent=0pt]
\setlength\itemsep{-0.25em}
    \item From $x_1 \cup x_2$:
        \begin{itemize}
        \setlength\itemsep{-0.1em}
            \item Subsequent Event: 80\% of the targets.
            \item Both Cause and Prerequisite: 60\% of the targets.
            \item Exclusively Cause: 28\% of the targets.
            \item Exclusively Prerequisite: 12\% of the targets.
        \end{itemize}
    \item From $x_2$: Motivation for all targets.
    \item From $x_3$: Reaction of listener for all targets.
\end{itemize}
\vspace{-\topsep}

\begin{table*}[t]
  \centering
  \begin{subtable}{\textwidth}
  \centering
  \resizebox{\textwidth}{!}{
    \begin{tabular}{p{\textwidth}}
    \toprule
    \textbf{A \pmb {($u_1$)}}: Hi, Jenny. Is it true you're moving to London?
    \textbf{B \pmb {($u_2$)}}: Yes, it is.
    \textbf{A \pmb {($u_3$)}}: What made you decide to do that?
    \textbf{B \pmb {($u_4$)}}: Work, mainly. I'm sure I'll be able to find a job there.
    \textbf{A \pmb {($u_5$)}}: You're probably right. But where are you going to live?
    \textbf{B \pmb {($u_6$)}}: I hope I'll find a flat to share with somebody. That way it will be cheaper.
    \textbf{A \pmb {($u_7$)}}: Yes, that's a good idea. Are you taking your dog with you?
    \textbf{B \pmb {($u_8$)}}:  No, I don't think so. My parents have offered to take care of him, and I don't think he'd be happy in the city.
    \textbf{A \pmb {($u_9$)}}: You're probably right. But aren't you afraid of moving to such a big place, especially after living in a small village?
    \textbf{B \pmb {($u_{10}$)}}: Not really. I think I'll enjoy myself. There's so much to do there; I expect I won't miss the countryside much and I can always come back and visit. 
    \textbf{A \pmb {($u_{11}$)}}: Well, I just hope you'll invite me to stay when you get settled.
    \textbf{B \pmb {($u_{12}$)}}:  Of course I will.
    \\
    \end{tabular}
    }
\end{subtable}
\bigskip
\begin{subtable}{\textwidth}
\centering
\resizebox{\textwidth}{!}{
  \begin{tabular}{p{\textwidth}}
      \toprule
      \textbf{Target -} \pmb{$u_6$}; \textbf{Inference:} \colorbox{almond}{\bf Cause}; \textbf{Annotation:}  Being an expensive city, it is quite difficult to find an affordable place to live in London. \\
      \midrule
      \textbf{Target -} \pmb{$u_{10}$}; \textbf{Inference:} \colorbox{almond}{\bf Cause}; \textbf{Annotation:} Jinny realizes that a city like London will provide a great quality of life for her. \\
      
      \midrule
      \textbf{Target -} \pmb{$u_6$}; \textbf{Inference:} \colorbox{ghostwhite}{\bf Subsequent Event}; \textbf{Annotation:}  The listener gives an idea to Jenny to find the flat on some online portal for searching flatmates as well plenty of cheaper options. \\
      
      \midrule
      \textbf{Target -} \pmb{$u_{10}$}; \textbf{Inference:} \colorbox{ghostwhite}{\bf Subsequent Event}; \textbf{Annotation:}  Jenny inquired a social club in London and ask for their membership to utilize her free time. \\
      
      \midrule
      \textbf{Target -} \pmb{$u_4$}; \textbf{Inference:} \colorbox{Green2}{\bf Prerequisite}; \textbf{Annotation:}  Jenny has completed her studies. \\
      \midrule
      \textbf{Target -} \pmb{$u_{12}$}; \textbf{Inference:} \colorbox{Green2}{\bf Prerequisite}; \textbf{Annotation:}  Jenny and the listener are good friends. \\
      \midrule
      \textbf{Target -} \pmb{$u_6$}; \textbf{Inference:} \colorbox{Blue2}{\bf Motivation}; \textbf{Annotation:}  Jenny is optimistic about having someone as her flatmate to save rent. \\
      
      \midrule
      \textbf{Target -} \pmb{$u_{12}$}; \textbf{Inference:} \colorbox{Gray3}{\bf Reaction}; \textbf{Annotation:}  The listener is happy for Jenny and looks forward to being invited to London by Jenny. \\
      \bottomrule
  \end{tabular}
  }
\end{subtable}
\caption{Annotated examples in \dataset{} marked with the target utterance and the inference type. Inference types \textit{Cause, Effect, Prerequisite, Motivation, and Reaction} correspond to questions Q1, Q2, Q3, Q4, and Q5, respectively, in \cref{sec:questions}.}
\label{tab:examples2}
\end{table*}

\subsection{Quality Assurance of \dataset{}}
Dataset quality is ensured with the following steps:
\begin{itemize}[leftmargin=*, wide, itemsep=0em, labelwidth=!, labelindent=0pt]
\setlength\itemsep{-0.25em}
\item Initially, we sample $50$ random dialogues and manually annotate all the questions (as in \cref{sec:questions}) in those. Each annotator is then evaluated on those dialogues, and is selected for the annotation task if 95\% of his/her annotations are approved by us.  
\item We constantly review and provide feedback to the annotators during the annotation process. Annotators are also instructed to amend their answers.
\item Upon completion of the annotation, we employ three additional annotators who manually check the annotated samples and score  their acceptability. These annotators reached a consensus for approving 86\% of these samples. The samples not bearing majority agreement were removed from the dataset. The statistics of the annotated dataset is shown in \cref{tab:stat}. A number of annotated examples from \dataset{} are also shown in \cref{tab:examples2}.
\end{itemize}

\subsection{Features of \dataset{}}
\label{sec:feat}

\begin{table}[t]
\centering
\resizebox{\linewidth}{!}{
	\begin{tabular}{p{4.5cm}@{}|c@{~~}|c@{~~}}
	\toprule
	\textbf{Description} & \textbf{\# Instances} & \textbf{Percentage}\\
	\midrule
	\bf \# Dialogues / \# Inferences & & \\
    $\quad$ DailyDialog & 3,280 / 30,509 & 57.82 / 57.34 \\
    $\quad$ MuTual & 1,640 / 14,207 & 28.91 / 26.70 \\
    $\quad$ DREAM & 753 / 8,488 & 13.27 / 15.95\\
    $\quad$ \bf Total & 5,673 / 53,204  & -- \\
    \midrule
    \# \bf Dialogues with \# Inferences & & \\
    $\quad$ less than 10 & 3,140 & 55.35 \\
    $\quad$ between 10-20 & 2,518 & 44.39 \\
    $\quad$ between 21-30 & 15 & 0.26\\
    \bf Avg. \# Inferences per Dialogue & 9.38 & --\\
    \midrule
    \begin{tabular}{l}
    \bf Instances with\\ \bf \# Correct Answers \end{tabular} & & \\
    $\quad$ only 1 & 45759 & 86.01 \\
    $\quad$ only 2 & 4985 & 9.37 \\
    $\quad$ $>$ 2 & 2460 & 4.62 \\
    \midrule
    \begin{tabular}{l}
    \bf Inference Types in \\ \bf Train / Validation / Test \end{tabular} & &  \\
    $\quad$ Cause & 10,386 / 3,060 / 3,071 & 33.06 / 28.10 / 28.18 \\
    $\quad$ Subsequent Event & 6,617 / 4,021 / 4,050 & 21.06 / 36.93 / 37.16 \\
    $\quad$ Prerequisite & 7,501 / 1,347 / 1,396 & 23.87 / 12.37 / 12.81 \\
    $\quad$ Motivation & 4,412 / 1,420 / 1,401 & 14.04 / 13.04 / 12.86 \\
    $\quad$ Reaction & 2,502 / 1,040 / 980\hspace{0.2cm} & 7.96\hspace{0.1cm}
    /\hspace{0.17cm}9.55\hspace{0.17cm}/\hspace{0.1cm} 8.99\\
    \bottomrule
	\end{tabular}
	}
	\caption{Statistics of the annotated \dataset{} dataset.}
	\label{tab:stat}
\end{table}

Following \cref{tab:stat}, a majority ($\sim 59\%$) of the inferences in \dataset{} are causal in nature.
Again, roughly 80\% of the inferences are speculative and context consonant. \dataset{} is thus much more versatile in terms of its applications as compared to CIDER~\cite{ghosal-etal-2021-cider} that only contains explicit contextual inferences. \dataset{} also contains varied commonsense knowledge -- from general to physical and social commonsense (see \cref{sec:appendix-ex} for more details).

\section{Commonsense Inference on \dataset{}}
\label{sec:tasks}

We design generative and multi-choice question answering tasks on \dataset{} to evaluate dialogue-level commonsense-based reasoning capabilities of language models.

\subsection{Task 1: \dataset{}$_{NLG}$}
The objective is to generate the answer to question $q$, representing one of the five inference types, for a target utterance $u_t$ in a dialogue $D$. Each inference type has its respective $q$ (illustrated in \Cref{sec:exp}).

\paragraph{Task 1.1: Dialogue Causal Inference.}
Causality pertains to causes and effects of events and situations. We formulate the dialogue causal inference task as generating the cause or subsequent event of an utterance as an answer to a causal question:

\begin{enumerate}[leftmargin=*, wide, itemsep=0em, labelwidth=!, labelindent=0pt, topsep=0cm]
    \item \textbf{Cause:} Given $D$, $u_t$, generate the cause $c_t$ of $u_t$. 
    \item \textbf{Subsequent Event:} Given $D$, $u_t$, generate the subsequent event $e_t$ of $u_t$. 
    \item \textbf{Subsequent Event Clipped (Subsequent EC):} Given $u_t$, the dialogue up to $u_t$: $D_{:u_t}$, generate the subsequent event $e_t$ of $u_t$. 
\end{enumerate}

We consider two  different scenarios for \textit{subsequent event}, as the event often appear after the target utterance in the dialogue. Hence, subtask 3 is more challenging to evaluate a model's ability to reason about unobserved effects. We extend subtasks 1, 2 to incorporate longer chains and formulate the chained generation task. We consider utterances $u_t$ in our dataset that has both cause and subsequent event annotated i.e. $c_t \rightarrow u_t \rightarrow e_t$. The causal chain is considered as a triplet, and we formulate tasks where a missing segment has to be generated from the rest of the components:

\begin{enumerate}[leftmargin=*, wide, itemsep=0em, labelwidth=!, labelindent=0pt, topsep=0cm]
\setcounter{enumi}{3}
    \item  \textbf{Chained Cause}: Generate $c_t$ from $u_t$ and $e_t$.
    \item  \textbf{Chained Subsequent Event (Chained SE)}: Generate $e_t$ from $u_t$ and $c_t$.
\end{enumerate}

\paragraph{Task 1.2: Prerequisite, Motivation and Reaction Generation.}
The objective is to generate the prerequisite/motivation/reaction of listener from a given $D$ and $u_t$. The target $u_t$ is the final utterance of $D$ for reaction generation. Generating the prerequisite (task 1.2.1) requires an understanding of the dependency of events. Generating the motivation (task 1.2.2) and reaction (task 1.2.3) is about learning basic human drives and emotions. Note that, reaction generation is a different problem from dialogue response generation. Responses follow utterance level distributions which are substantially different from emotional reactions.

\subsection{Task 2: \dataset{}$_{MCQ}$}
\label{sec:altsc}
\begin{figure*}[t]
    \centering
        \includegraphics[width=0.8\linewidth]{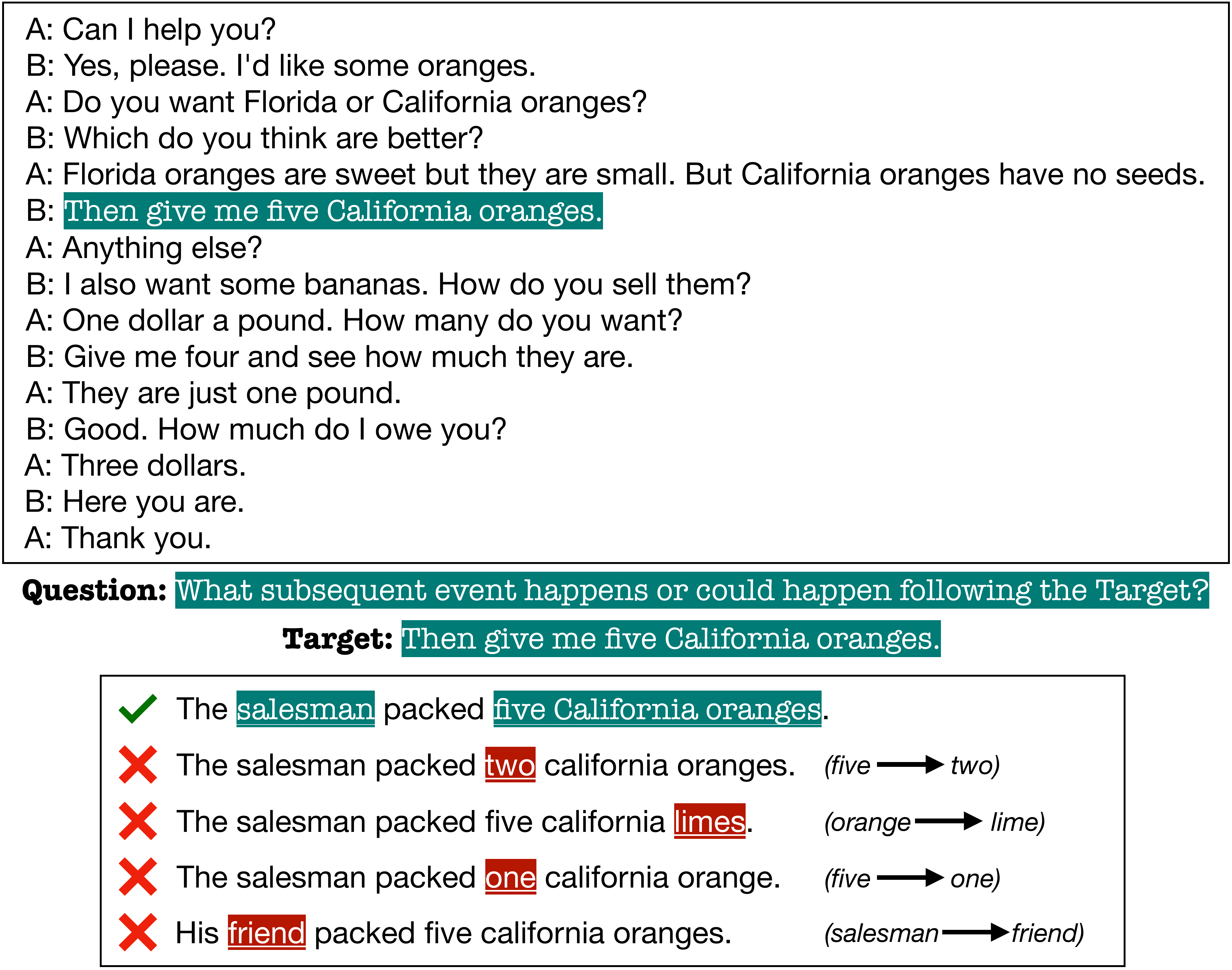}
        \caption{A data sample of \dataset{} for the Plausible Alternative Selection task. Here, commonsense is required to infer -- a salesman packs the items that buyers want to purchase. In this particular dialogue, the buyer wants to purchase five California oranges and four bananas which can be inferred from the context.}
        \label{fig:mcq}
\end{figure*}

Given dialogue $D$, target $u_t$, one of the five questions (inference type) $q$, true answer $a_t$, alternate choices $F_t = \{f_{t1}, f_{t2}, f_{t3}, f_{t4}\}$, the \dataset{}$_{MCQ}$ task aims to select the correct answer $a_t$ (see \cref{fig:mcq}) and additionally any answer among $F_{t}$ which might be correct. The alternate choices $F_{t}$ are created through a combination of automated generation and human supervision as follows:

\begin{itemize}[leftmargin=*, wide, itemsep=0em, labelwidth=!, labelindent=0pt, topsep=0cm]
\item We train a \code{T5} large model on SNLI contradictory pairs~\cite{bowman2015large} and Time-Travel counterfactual pairs~\cite{qin2019counterfactual} to generate contradictions/counterfactuals from input sentences. We use this model to generate a pool of alternate answers from the true annotated answers. Alternate answers which have an embedding cosine similarity less than 0.9 with the true answer (from \textit{all-mpnet-base-v2} in ~\citet{reimers-2019-sentence-bert}) and are contradictory w.r.t the true answer (from \textit{roberta-large-mnli}) are kept, and the rest are discarded. The filtered set is termed $\mathbb{N}$.

\item We use the adversarial filtering (AF) algorithm~\cite{zellers2018swag} to select the four alternate answers $F_{t}$ from $\mathbb{N}$. For multi-choice QA tasks, AF is an effective method to detect easily identifiable alternate answers and replace them with more difficult candidates by detecting and reducing stylistic artifacts. The algorithm is as follows:

\begin{enumerate}[label=\bfseries(\roman*), leftmargin=*, wide, itemsep=0em, labelwidth=!, labelindent=0pt, topsep=-0.01cm]
\item We start with annotated true answer $a_t$ and any four choices $\hat{F_{t}}$ from $\mathbb{N}$ for all instances in our dataset to create $\hat{\mathbb{D}}$. We randomly split $\hat{\mathbb{D}}$ into $\hat{\mathbb{D}}_{train}$ (80\%) and $\hat{\mathbb{D}}_{test}$ (20\%) according to dialogue IDs. 
\item A multi-choice QA model (discriminator) is trained on $\hat{\mathbb{D}}_{train}$ that scores all five choices for all instances in $\hat{\mathbb{D}}_{test}$. The highest scoring choice is considered as the predicted answer. For a particular test instance, choices in $\hat{F_{t}}$ that have lower scores than $a_t$ are replaced with other high scoring choices in $\mathbb{N} - \hat{F_{t}}$. Answers in $\hat{F_{t}}$ which are being replaced are removed from $\mathbb{N}$.  
\item $\hat{F_{t}}$ now consists of relatively more difficult choices. A new random split $\hat{\mathbb{D}}_{train}$ and $\hat{\mathbb{D}}_{test}$ is created, and we go back to step \textbf{(ii)}. The algorithm is terminated when the accuracy in successive $\hat{\mathbb{D}}_{test}$ reaches a convergence. The final alternate choice set is termed as $F_t$.
\end{enumerate}
The AF algorithm ensures a robust final dataset $\mathbb{D}$ irrespective of the final train, validation, and test split. We use a new \textit{roberta-large} model to initialize the discriminator and train for 3 epochs before scoring and replacement in step \textbf{(ii)}. 14 iterations were required for convergence in $\mathbb{D}_{test}$.

\item Annotators perform manual checking on the final AF selected choices $F_t$. They mark each of the alternate choices in $F_{t}$ in $\mathbb{D}$ to be speculatively correct or incorrect given the context. Hence, instances might have correct answers in $F_{t}$ in addition to the originally annotated correct answer $a_t$. The final dataset statistics after this step are given in \cref{tab:stat}.
\end{itemize}

\paragraph{Task 2.1: Single Answer Selection.}
Consider instances where $F_t$ doesn't contain any correct answer. The task is to select the correct answer $a_t$ among the five choices given $D$, $u_t$, and $q$. 

\paragraph{Task 2.2: All Answers Selection.}
This task is performed on the entire dataset (including the subset of data which is used in Task 2.1. There might be one or more correct answers for a particular instance resulting from the AF algorithm. The task is to select all the correct answer(s) (including $a_t$) among the five choices given $D$, $u_t$, and $q$.

\section{\dataset{} Tasks: Experimental Results}
\label{sec:exp}
We split our dataset in dialogue level where the training, validation and test instances are obtained from a total of 3477, 1097, 1098 distinct dialogues respectively. This results in a 60:20:20 proportion of total annotation instances. The three sets have 17365, 5370, and 5331 unique target utterances respectively. We tune on the validation dataset and report results on the test dataset (average of 5 runs). For the sake of brevity, the detailed hyperparameters are given in the supplementary material.

We use the following questions ($q$) for the five inference types for all the tasks: 
\textbf{Cause}: \textit{What is or could be the cause of target?}
\textbf{Subsequent Event}: \textit{What subsequent event happens or could happen following the target?}
\textbf{Prerequisite}: \textit{What is or could be the prerequisite of target?}
\textbf{Motivation}: \textit{What is or could be the motivation of target?}
\textbf{Reaction}: \textit{What is the possible emotional reaction of the listener in response to target?}

\subsection{Baseline Models}
\label{sec:baselines}
\paragraph{\dataset{}$_{NLG}$ --- (1.1--1.2).} We use large versions of \code{T5}~\cite{raffel2019exploring} and \code{GLUCOSE-T5}~\cite{mostafazadeh-etal-2020-glucose} as our models. \code{GLUCOSE-T5} is a T5 large model that is pre-trained on the GLUCOSE dataset. We concatenate $q$, $u_t$, and the context $c$ with separators to form the input to the model: \code{q <sep> $u_t$ <sep> c}. The context $c$ is formed by concatenating utterances of $D_{:u_t}$ (subsequent event clipped) or $D$ (all other tasks).
For the chained generation task, we additionally provide the cause/subsequent event as input. The inputs are \code{q <sep> $u_t$ <sep> subsequent event: $e_t$ <sep> c} and \code{q <sep> $u_t$ <sep> cause: $c_t$ <sep> c} for cause and subsequent event generation, respectively. The objective is to generate the answer as output in the sequence-to-sequence setup. We use teacher forcing during training and beam search during inference.

\paragraph{\dataset{}$_{MCQ}$ --- Single Answer Selection (2.1).}
We use \code{RoBERTa-large}, \code{ELECTRA-large}, \code{T5-large}, and \code{Unified QA Large} for this task. The input to the models for \code{RoBERTa-large}, \code{ELECTRA-large} is the concatenation of question $q$, target $u_t$, dialogue $D$, and candidate answers $x_j, j \in \{1,...,5\}$: \code{<cls> q <sep> $u_t$ <sep> D <sep> $x_j$}. Each score is predicted from the corresponding \code{<cls>} vector and the highest scoring one is selected as the answer. For seq2seq models \code{T5-large}, and \code{Unified QA Large}, we use the following input --- \code{q <sep> 1) $x_1$ 2) $x_2$ 3) $x_3$ 4) $x_4$ 5) $x_5$ <sep> $u_t$ <sep> D}. The output to be generated is the correct answer -- such as \code{$x_1$} or \code{$x_2$}.

\paragraph{\dataset{}$_{MCQ}$ --- All Answers Selection (2.2).}
We use seq2seq models \code{T5-large}, and \code{Unified QA Large} as they can generate both single and multiple-answers (with separator tokens) as output. The input is \code{q <sep> 1) $x_1$ 2) $x_2$ 3) $x_3$ 4) $x_4$ 5) $x_5$ <sep> $u_t$ <sep> D}. The output to be generated are the correct answer(s), such as \code{$x_2$} (single answer) or \code{$x_1$ <sep> $x_3$ <sep> $x_4$} (multiple answers). Here, $x_1 - x_5$ denotes the five possible choices shuffled randomly. 

\fboxsep0pt
\begin{table*}[t]
\centering
\resizebox{0.8\textwidth}{!}{
\begin{tabular}{ll|cccccccc}
\toprule
&\textbf{Model} & \textbf{BLEU2} & \textbf{METEOR} & \textbf{ROUGE} & \textbf{CIDEr} & \textbf{Sem-Sim} \\
\midrule

\rotatetabular{4}{almond}{(1.1.1)}{Cause}& 
\code{T5}  & 	 0.1493  &  	 0.1630  & 	 0.2626  & 	0.4560  & 	 0.6278  \\
 & \code{GLUCOSE-T5}  &	\bf 0.1563 &	\bf 0.1634 &	\bf 0.2707 &	\bf 0.4915 &	\bf 0.6305 \\
 & \code{T5$^*$} &	0.0042 & 0.0200 &	0.0266	& 0.0237 &	0.3735 \\
 & \code{GLUCOSE-T5$^*$} &	0.0287 &	0.0560 &	0.0827 &	0.1332 &	0.4442 \\
 \cmidrule{2-7}

\rotatetabular{4}{ghostwhite}{(1.1.2)}{SE}& 
\code{T5}  & 	 \textbf{0.1619}  & 	 \textbf{0.1662}  & 	 0.2760  & 	 0.4119  & 		 0.6276 \\
 & \code{GLUCOSE-T5}  &	0.1611 &	0.1628 &	\textbf{0.2778} &	\textbf{0.4430} &	\textbf{0.6297} \\
 & \code{T5$^*$} &	0.0045 &	0.0191 &	0.0264 &	0.0241 &	0.3865 \\
 & \code{GLUCOSE-T5$^*$}  &	0.0001 &	0.0070 &	0.0024 &	0.0032 &	0.3073\\
 
 \cmidrule{2-7}

\rotatetabular{4}{ghostwhite}{(1.1.3)}{SE Clipped}& 
\code{T5}   & 	 0.1448  & 	 \textbf{0.1549}  & 	 0.2618  & 	 0.3099  & 	\textbf{0.6123} \\
 & \code{GLUCOSE-T5}  &	\textbf{0.1461}  &	0.1523  &	\textbf{0.2645}  &	\textbf{0.3238}  &	0.6094\\
 & \code{T5$^*$} &	0.0199  &	0.0439  &	0.0564  &	0.0762  &	0.4549  \\
 & \code{GLUCOSE-T5$^*$}  &	0.0001  &	0.0066  &	0.0025  &	0.0034  &	0.3063\\

\cmidrule{2-7}

\rotatetabular{4}{Green2}{(1.2.1)}{Prerequisite}& 
\code{T5}   & 	 \textbf{0.1002}  & 	 0.1282  & 	 0.2176  & 	 \textbf{0.3357}  &  	\textbf{ 0.5902} \\
 & \code{GLUCOSE-T5}  &	0.1001 &	\textbf{0.1299} &	\textbf{0.2197} &	0.3144 &	0.5896 \\
 & \code{T5$^*$} &	0.0043 &	0.0222 &	0.0279 &	0.0225 &	0.3541 \\
 & \code{GLUCOSE-T5$^*$}  &	0.0108 &	0.0394 &	0.0625 &	0.0889 &	0.4392\\

\cmidrule{2-7}

\rotatetabular{4}{Blue2}{(1.2.2)}{Motivation}&
\code{T5}   & 	 0.2503  & 	 0.1998  & 	 0.3781  & 	 0.7109  & 		 0.6973 \\
 & \code{GLUCOSE-T5}  &	\textbf{0.2582} &	\textbf{0.2037} &	\textbf{0.3840} &	\textbf{0.7499} &	\textbf{0.7048} \\
 & \code{T5$^*$} &	0.0033 & 0.0183 &	0.0257 &	0.0181 &	0.4038 \\
 & \code{GLUCOSE-T5$^*$} &	0.0174 &	0.0434 &	0.0632 &	0.0696 &	0.4053\\

\cmidrule{2-7}

\rotatetabular{4}{Gray3}{(1.2.3)}{Reaction}&
 \code{T5}  &  	 \textbf{0.2397}  &  	 \textbf{0.1939}  & 	 \textbf{0.3720}  & 	 0.5177  & 	 \textbf{0.6665}  \\
 & \code{GLUCOSE-T5}  &	0.2318 &	0.1903 &	0.3716 &	\textbf{0.5364} &	0.6653\\
 & \code{T5$^*$}  &	0.0037 &	0.0201 &	0.0239 &	0.0167 &	0.3899 \\
 & \code{GLUCOSE-T5$^*$}  &	0.0213 &	0.0459 &	0.0759 &	0.0719 &	0.4125\\
 
\cmidrule{1-7}
\rotatetabularnormal{1}{brilliantlavender}{Average}{Score}&
 \code{T5}  & 0.1744 & \textbf{0.1677} & 0.2947 & 0.4570 & 0.6369 \\
 & \code{GLUCOSE-T5}  & \textbf{0.1756} & 0.1671 & \textbf{0.2980} & \textbf{0.4765} & \textbf{0.6382} \\
 & \code{T5$^*$}  & 0.0066 & 0.0239 & 0.0312 & 0.0302 & 0.3938 \\
 & \code{GLUCOSE-T5$^*$}  & 0.0130 & 0.0322 & 0.0491 & 0.0601 & 0.3886 \\

\bottomrule
\end{tabular}
}
\caption{Results of the \dataset{}$_{NLG}$ task. \code{T5$^*$} and \code{GLUCOSE-T5$^*$} are not fine-tuned on our dataset. All models are Large models. \colorbox{ghostwhite}{SE} denotes Subsequent Event.}
\label{tab:results}
\end{table*}

\begin{table*}[t]
\centering
\resizebox{0.8\linewidth}{!}{
\begin{tabular}{l|ccccc}
\toprule
\textbf{Model} & \textbf{BLEU2} & \textbf{METEOR} & \textbf{ROUGE} & \textbf{CIDEr} & \textbf{Sem-Sim} \\
\midrule
\colorbox{almond}{\textbf{(1.1.4) Chained Cause}} & & \\
\quad \quad \code{T5}    & 	 0.1566   & 	0.1675  & 	 0.2757  & 	 0.5303  & 	 0.6518 \\
\quad \quad \code{GLUCOSE-T5}    &	0.1600  &	\bf 0.1697  &	\bf 0.2796  &	\bf 0.5633  &	\bf 0.6557\\
\cmidrule{2-6}
\colorbox{almond}{\textbf{(1.1.1)* Cause}} & & \\
\quad \quad \code{T5}   & 	 0.1503   & 	 0.1635  & 	 0.2634  &  0.4591  & 0.6284  \\
\quad \quad \code{GLUCOSE-T5}     &	\bf 0.1564  &	\bf 0.1636  &	\bf 0.2709  &	\bf 0.4915  &	\bf 0.6310\\
\midrule
\colorbox{ghostwhite}{\textbf{(1.1.5) Chained SE}} & & \\
\quad \quad \code{T5}   & 	 \textbf{0.1813}   & 	 \textbf{0.1784}  & 	 0.2940  & 	 0.5136  & 	 0.6469 \\
\quad \quad \code{GLUCOSE-T5}    &	0.1789  &	0.1776  &	\bf 0.2943  &	\bf 0.5218  &	\bf 0.6516\\
\cmidrule{2-6}
\colorbox{ghostwhite}{\textbf{(1.1.2)* SE}} & & \\
\quad \quad \code{T5}   & 	 \bf 0.1622  & 	 0.0841   & 	 0.2764  & 	 0.4167  & 0.6279  \\
\quad \quad \code{GLUCOSE-T5}    &	0.1612  &	\bf 0.1628  &	\bf 0.2778  &	\bf 0.4471  &	\bf 0.6294\\
\bottomrule
\end{tabular}
}
\caption{Results of the \dataset{}$_{NLG}$ subtasks -- chained cause and subsequent event generation. (1.1.1)* and (1.1.2)* indicates results from Task 1.1.1 and 1.1.2 (as in \cref{tab:results}), but only for targets which have both cause and effect annotated, ensuring a fair comparison with (1.1.4) and (1.1.5). \colorbox{ghostwhite}{SE} denotes Subsequent Event.}
\label{tab:cqa}
\end{table*}

\subsection{Results of the \dataset{}$_{NLG}$ Task}
\label{sec:results}

\paragraph{Automatic Evaluation Metrics.} For generative tasks, we report the following metrics: \textbf{BLEU}~\cite{papineni2002bleu}, \textbf{METEOR}~\cite{banerjee2005meteor}, \textbf{ROUGE}~\cite{lin2004rouge}, \textbf{CIDEr}~\cite{vedantam2015cider}, and \textbf{Sem-Sim} which computes the semantic cosine similarity of two sentences using the supervised \code{RoBERTa-large} sentence embedding model ~\cite{gao2021simcse}. All scores are reported in the range of 0-1.

\paragraph{Human Evaluation Metrics.} Due to significant dissonance with human evaluation, automatic evaluation metrics are often considered not reliable for generation quality evaluation in literature. Hence, we resort to human evaluation metrics. The human annotators rate on an integer scale from 1 (worst) to 5 (best) on three coarse attributes:
\textbf{Creativity}: As the majority of the inferences require speculation, this metric measures how creative the models and the annotators are. \textbf{Contextuality}: Whether the generated or annotated inferences fit the context. \textbf{Fluency}: Whether the generated or annotated inferences are grammatically correct.

\paragraph{Results of Automatic Evaluation.} The results for the generative tasks are reported in \cref{tab:results} and \cref{tab:cqa}. We observe that the fine-tuned models perform quite similarly across various metrics in \cref{tab:results}. The \code{T5} model achieves the best performance in most of the experimental settings.
The results indicate that the \emph{causal} types are more challenging to infer than the \emph{Motivation}, and \emph{Reaction}. However, the models are posed to the most challenging instances in the case of \emph{Prerequisite} type as inferring this type requires rich commonsense and background knowledge. Hence, for this category, the models achieve a low score compared to rest of the inference categories. We also notice that exposing the future utterances to the models help in attaining better inference performance for the relation type \emph{Subsequent Event}. The trained models perform worse when the future utterances are not available in the input as seen in the \emph{Subsequent Event Clipped} task. A significant drop of performance is noticed in the CIDEr metric. For the chained generation tasks (1.1.4 and 1.1.5), we notice (refer to \cref{tab:cqa}) a very similar trend in models' performance i.e., the models tend to perform better for these two experimental settings compared to only \emph{Cause} (1.1.1) and \emph{Subsequent Event} (1.1.2) predictions. We can surmise that the additional cues from the available annotations of \emph{Subsequent Event} type in the Chained Cause setting, and the \emph{Cause} type in the Chained Subsequent Event setting are the key to such performance improvement. As depicted in \cref{tab:results} (and also \cref{tab:w/o-fine-tune}), the non fine-tuned versions of \code{T5} and \code{GLUCOSE-T5} perform poorly as they produce gibberish outputs across all the five inference categories indicating the importance of fine-tuning on \dataset{}.

\begin{table}[t]
\centering
\resizebox{\linewidth}{!}{
\begin{tabular}{l|ccccccccc}
\toprule
\bf Metric        & \bf Gold & \bf \code{T5}  & \bf GLUCOSE & \bf T5$^*$ & \bf GLUCOSE$^*$ \\
\midrule
Creativity    & 4.7  & 3.8 & 3.9  & 2.4  & 1.9   \\
Contextuality & 4.8  & 4.1 & 4.3  & 2.1  & 2.1  \\
Fluency       & 5.0  & 4.8 & 4.9  & 1.9  & 2.9  \\
\bottomrule
\end{tabular}
}
\caption{Results of the human evaluation for \dataset{}$_{NLG}$. \code{T5$^*$} and \code{GLUCOSE-T5$^*$} represent non fine-tuned versions.}
\label{tab:w/o-fine-tune}
\end{table}
\begin{table*}[t]
  \centering
  \begin{subtable}{\textwidth}
  \centering
  \resizebox{\textwidth}{!}{
    \begin{tabular}{p{\textwidth}}
    \toprule
    \textbf{A \pmb {($u_1$)}}: I'm hungry, let's order up something to eat.
    \textbf{B \pmb {($u_2$)}}:  Ok, maybe we can order a soup and a salad from the restaurant down the street.
    \textbf{A \pmb {($u_3$)}}: I was thinking of getting a hamburger, fries and a chocolate sundae.
    \textbf{B \pmb {($u_4$)}}: You eat too much junk food. That sort of stuff clogs up your arteries and is very high in cholesterol.
    \textbf{A \pmb {($u_5$)}}: Well I never seem to gain weight so I don't mind.
    \textbf{B \pmb {($u_6$)}}: It's not only about getting fat or not, it's about being healthy. You could really have some health problems later on.
    \textbf{A \pmb {($u_7$)}}: How about pizza or maybe some fried chicken! Better yet, let's order some hot dogs!
    \textbf{B \pmb {($u_8$)}}:  You are a lost cause.\\
    \end{tabular}
    }
\end{subtable}
\bigskip
\begin{subtable}{\textwidth}
\centering
\resizebox{\textwidth}{!}{
  \begin{tabular}{p{\textwidth}}
      \toprule
      \textbf{Target -} \pmb{$u_1$}; \textbf{Inference:} \colorbox{almond}{\bf Cause}; \textbf{Answers:} \colorbox{gold}{\textbf{Gold}} - The speaker did not eat breakfast. \colorbox{Blue4}{\textbf{T5}} - The speaker and the listener are going to a restaurant for lunch. 
      \colorbox{aqua}{\textbf{GLUCOSE-T5}} - The speaker and the listener are at a restaurant. \\
      \midrule
      \textbf{Target -} \pmb{$u_2$}; \textbf{Inference:} \colorbox{almond}{\bf Cause}; \textbf{Answers:} \colorbox{gold}{\textbf{Gold}}  - The restaurant down the street is famous for its delicious food. \colorbox{Blue4}{\textbf{T5}} - The listener is hungry and wants to eat something.
      \colorbox{aqua}{\textbf{GLUCOSE-T5}} - The speaker is hungry and wants to order something to eat. \\
      \midrule
      \textbf{Target -} \pmb{$u_3$}; \textbf{Inference:} \colorbox{Blue2}{\bf Motivation}; \textbf{Answers:} \colorbox{gold}{\textbf{Gold}} - The speaker is craving some savory fast food. \colorbox{Blue4}{\textbf{T5}} - The speaker is craving for a hamburger, fries and a chocolate sundae.
      \colorbox{aqua}{\textbf{GLUCOSE-T5}} - The speaker is craving for a burger, fries and sundae. \\
      \midrule
      \textbf{Target -} \pmb{$u_6$}; \textbf{Inference:} \colorbox{Green2}{\bf Prerequisite}; \textbf{Answers:} \colorbox{gold}{\textbf{Gold}} - The speaker is a fitness freak and keeps track of his daily diet. \colorbox{Blue4}{\textbf{T5}} - The speaker is a healthy person.
      \colorbox{aqua}{\textbf{GLUCOSE-T5}} - The speaker is a health conscious person. \\
      \midrule
      \textbf{Target -} \pmb{$u_7$}; \textbf{Inference:} \colorbox{ghostwhite}{\bf Subsequent Event}; \textbf{Answers:} \colorbox{gold}{\textbf{Gold}} - The listener refused to eat anything that is unhealthy. \colorbox{Blue4}{\textbf{T5}} - The speaker and the listener decided to order some hot dogs.
      \colorbox{aqua}{\textbf{GLUCOSE-T5}} - The speaker and the listener decided to order some hot dogs. \\
      \midrule
      \textbf{Target -} \pmb{$u_8$}; \textbf{Inference:} \colorbox{Gray3}{\bf Reaction}; \textbf{Answers:} \colorbox{gold}{\textbf{Gold}} - The listener felt embarrassed by the statement of the speaker. \colorbox{Blue4}{\textbf{T5}} - The listener is shocked to hear the speaker’s comment.
      \colorbox{aqua}{\textbf{GLUCOSE-T5}} - The listener is disappointed with the speaker's decision. \\
      \bottomrule
  \end{tabular}
  }
\end{subtable}
\caption{Inferences by different models extracted from a sample dialogue for the \dataset{}$_{NLG}$ task.}
\label{tab:examples}
\end{table*}

\paragraph{Results of Human Evaluation.} For each of the five inference types, we randomly sample $40$ inferences generated by each model and their corresponding gold inferences. These inferences are then manually rated by three independent annotators based on the human-evaluated metrics. As suggested by \cref{tab:w/o-fine-tune}, we observe that most of the fine-tuned models on \dataset{} perform similarly but fail to reach gold annotation performance. Moreover, as expected, the fine-tuned models significantly outperform their non fine-tuned counterparts. We provide some examples of the generated inferences in \cref{tab:examples}. Inspection of the model generated inferences reveal that usage of keywords from the dialogue without generalizing the events is more frequent. Generated inferences are significantly less diverse and creative than gold annotations. 

\paragraph{Performance of GLUCOSE.} GLUCOSE contains contextual commonsense inferences on events in monologues. Comparing the results (\cref{tab:results}, \cref{tab:w/o-fine-tune}) of fine-tuned and non fine-tuned checkpoints suggests that pre-training on a monologue-based contextual commonsense inference dataset does not ensure good performance on the same task for dialogues. Akin to the non fine-tuned T5, non fine-tuned \code{GLUCOSE-T5} produces gibberish outputs for all the commonsense inference types but the causal and motivation types. We surmise this happens as these two commonsense types exist in the GLUCOSE dataset. Although the generated text for these two commonsense inference types are grammatically correct and sometimes contain contextual words, they are far from the desired quality, semantically very much dissimilar from the annotated gold instances, and rated low in the qualitative evaluation, as shown in \cref{tab:w/o-fine-tune}. We also confirm the efficacy of fine-tuning the models on \dataset{} through human evaluation, as explained in \cref{sec:exp}. 

\begin{table*}[t]
\centering
\resizebox{0.9\linewidth}{!}{
\begin{tabular}{l|cccccc}
\toprule
\bf Model & \bf Cause & \bf SE & \bf Prerequisite & \bf Motivation & \bf \begin{tabular}[c]{@{}c@{}}Emotional\\ Reaction\end{tabular} & \bf Average \\
\midrule
\code{RoBERTa}  & 83.34 & 83.17 & 79.48 & 86.33 & 84.26 & 83.28 \\
\code{ELECTRA}  &  \textbf{87.09}  & \textbf{86.09} &  \textbf{85.15} & \textbf{90.31} & \textbf{86.11} & \textbf{86.82} \\
\cmidrule{2-7}
\code{T5} & 95.19 & \textbf{95.29} & 94.93 & \textbf{96.52} & 96.99 & 95.54 \\
\code{Unified QA}  & \textbf{95.85} & 94.99 & \textbf{95.55}  & 96.35 & \textbf{97.22} & \textbf{95.70}\\
\bottomrule
\end{tabular}
}
\caption{Accuracy scores for Task 2.1. Models are trained and evaluated on instances with a single correct answer.}
\label{tab:alternative}
\end{table*}

\begin{table*}[t]
\centering
\resizebox{0.9\linewidth}{!}{
\begin{tabular}{l|c|cccccc}
\toprule
\bf Model \bf & \bf \begin{tabular}[c]{@{}c@{}}Evaluated\\ On\end{tabular}  & \bf Cause & \bf SE & \bf Prerequisite & \bf Motivation & \bf \begin{tabular}[c]{@{}c@{}}Emotional\\ Reaction\end{tabular} & \bf Average \\
\midrule
\code{T5} & \multirow{2}{*}{S + M} &  \textbf{78.18} & 74.72 & \textbf{75.50} & \textbf{82.51} & \textbf{84.59} & \textbf{77.68} \\
\code{Unified QA}  & & 78.12 & \textbf{74.79} & 75.36 & 81.58 & 84.08 & 77.51\\
\midrule
\code{T5} & \multirow{2}{*}{S} &  \textbf{93.20} & \textbf{91.28} & \textbf{91.27} & \textbf{95.19} & \textbf{95.14} & \textbf{92.71} \\
\code{Unified QA}  &  & 93.12 & 91.16 & 91.00 & 94.28 & 94.79 & 92.45 \\
\midrule
\code{T5} & \multirow{2}{*}{M} & \bf 3.50 & 2.77 & \bf 3.59 & \bf 3.61 & \bf 6.03 & 3.38 \\
\code{Unified QA} &  & \bf 3.50 & \bf 3.69 & \bf 3.98 & 2.58 & \bf 4.31 & \bf 3.60 \\
\bottomrule
\end{tabular}
}
\caption{Exact match scores for Task 2.2. Models are trained on instances with both single and multiple correct answers, i.e., the entire dataset. SE $\rightarrow$ Subsequent Event; S $\rightarrow$ Single-Answer Instances; M $\rightarrow$ Multi-Answer Instances.}
\label{tab:alternative2}
\end{table*}

\subsection{Results of the \dataset{}$_{MCQ}$ Task}
\label{sec:results-mcq}
\paragraph{Evaluation Metrics.} 1) \textbf{\code{RoBERTa}} and \textbf{\code{ELECTRA}}: The accuracy of selecting the correct answer is used to evaluate the performance of these models.
2) \textbf{\code{T5}} and \textbf{\code{Unified QA}}: The output is considered as a single answer if it doesn't contain any separator token. Otherwise, the output is segmented at separator tokens to obtain multiple answers. We then follow the method in \citet{2020unifiedqa}, where match is computed by comparing each of the generated answer(s) with the candidate choices based on their token-level overlap. For each generated answer, the most similar candidate choice is considered as the corresponding output. The prediction is considered as correct if the final output(s) is an exact match (EM) with the gold annotated answer(s).

\paragraph{Single Answer Selection (2.1).}
We report the results of this setting in \cref{tab:alternative}. The reported metric is accuracy of selecting the correct answer. The overall score is 83.28\% for \code{RoBERTa} and 86.82\% for \code{ELECTRA}. \code{ELECTRA} has an edge over \code{RoBERTa} on all the five inference types. This could be a side effect of using \code{RoBERTa} as the backbone model for the AF algorithm and subsequently as a solver for the final \dataset{}$_{MCQ}$ task. We think, this results expose the model dependency of the AF process. In other words, the negative samples chosen by the backbone model $X$ for the AF algorithm will be difficult to distinguish from the human-annotated true samples using the same model $X$. These negative samples, however, could be relatively easier to identify using another model $Y$. The seq2seq models \code{T5} and \code{Unified QA} perform significantly better than \code{RoBERTa} and \code{ELECTRA} as can be seen in \cref{tab:alternative}. While models like \code{RoBERTa, ELECTRA} encode each candidate answer separately, \code{T5} and \code{Unified QA} encode them together. Thanks to this joint encoding of candidate answers, \code{T5} and \code{Unified QA} can take advantage of more task-related information that \code{RoBERTa} and \code{ELECTRA} might miss due to the separate encoding scheme. We surmise it could be one of the reasons why the seq2seq models have an edge over \code{RoBERTa} and \code{ELECTRA} for this particular task. \code{T5} and \code{Unified QA} attain almost the same score for single answer selection. This is surprising as \code{Unified QA} is initialized from the \code{T5-large} checkpoint and then further trained on other QA datasets. As such, we think, the different fine-tuned domains of \code{Unified QA} does not help in the \dataset{}$_{MCQ}$ task.

\paragraph{All Answers Selection (2.2).}
We train and evaluate \code{T5} and \code{Unified QA} on the entire dataset of both single and multiple correct answers and report the results in \cref{tab:alternative2}. Overall, \code{T5} and \code{Unified QA} perform similarly. 
The general performance, across the models, on instances with multiple correct answers is much worse than instances with a single correct answer. We confirm this by reporting the results only on instances with multiple answers in \cref{tab:alternative2}, where \code{T5} and \code{Unified QA} achieve only 3.38\% and 3.60\% exact match, respectively. This could probably be attributed to the stark data imbalance of $\sim$86/14\% between single- and multi-answer instances, respectively (see \cref{tab:stat}).

\section{Related Work}
\label{sec:rwork}
Commonsense knowledge has received more attention compared with factual knowledge, as it is usually not mentioned explicitly in the context. It is demonstrated to be essential in open-ended generation tasks, such as story explanation generation~\cite{mostafazadeh-etal-2020-glucose}, story end generation \cite{Guan2019StoryEG} and abductive reasoning~ \cite{bhagavatula2019abductive}. To infuse commonsense knowledge in NLP models, several approaches to tasks like sentence ordering~\cite{stack}, emotion recognition~\cite{cosmic}, story generation~\cite{Guan2020AKP,Xu2020ControllableSG} and dialogue generation~\cite{Zhou2018CommonsenseKA} use prevalent commonsense knowledge bases (CSKB) like ConceptNet~\cite{speer2017conceptnet} or ATOMIC \cite{sap2019atomic}. However, ConceptNet is context-free, meaning that they only capture relationships around a selected set of entities, without paying attention to the context where the entity occurs. Moreover, inference is often needed in discourse level, which do not always align with the entities in knowledge bases. Knowledge models such as COMET~\cite{ bosselut2019comet} is a way to circumvent this issue and make inferences on an utterance (sentence) level.  But the generated knowledge still lacks the detail from the dialogue, as it is trained on the aforementioned knowledge base. Our approach, instead, centers on the dialogue dataset and provides more detailed commonsense inference at an utterance level. 

\section{Conclusion}
We introduced \dataset{}, a new dataset for dialogue reasoning with contextualized commonsense inference. It contains $\sim$53K inferences for five commonsense dimensions -- cause, subsequent event, prerequisite, motivation, and emotional reaction -- collected from $\sim$5.6K dialogues. To show the usefulness of \dataset{} for dialogue reasoning, we design several challenging generative and multi-choice answer selection tasks for state-of-the-art NLP models to solve.

\section*{Acknowledgements}
This work is supported by the A*STAR under its RIE 2020 AME programmatic grant
RGAST2003 and project T2MOE2008 awarded by Singapore's MoE under its Tier-2 grant scheme.

\section*{Ethics Statement}
The annotators for \dataset{} were hired through a data annotation service. The compensation was derived based on the country of residence of the annotators, as deemed by the company. The study has been categorized as ``exempt'' by the IRB. Annotators were strictly asked not to write any toxic content (hateful or offensive toward any gender, race, sex, religion). They were asked to consider gender-neutral settings in dialogues whenever possible.

The source dialogue datasets -- DailyDialog, MuTual, and DREAM are high quality multi-turn dialogue datasets manually annotated by experts in dialogue, communication theory and linguistics. All three datasets have been extensively used and studied in the natural language processing literature. The three source datasets and our annotations in \dataset{} do not contain any personal data or any information that can uniquely identify individual people or groups.

\bibliography{refs}
\bibliographystyle{acl_natbib}
\clearpage

\appendix

\begin{table*}[h!]
\small
\centering
\resizebox{0.8\textwidth}{!}{
\begin{tabular}{ll|cccccccc}
\toprule
& \textbf{Model} & \textbf{BLEU1} & \textbf{BLEU2} & \textbf{METEOR} & \textbf{ROUGE} & \textbf{CIDEr} & \textbf{Sem-Sim} \\
\midrule

\rotatetabularnormal{8}{almond}{(1.1.1)}{Cause}& 
\code{T5} & 0.2874  & 	 0.1493  &  	 0.1630  & 	 0.2626  & 	0.4560  & 	 0.6278  \\
  &  \code{BART} &  0.2542  & 	 0.1396  & 	 0.1527  & 	 0.2586  & 	 0.4241  & 	 	 0.6224 \\
 &  \code{COMET} & 0.2762 &	0.1518 &	0.1580 &	0.2652 &	0.4486 &	0.6253 \\
 & \code{GLUCOSE-T5} & \bf 0.2935 &	\bf 0.1563 &	\bf 0.1634 &	\bf 0.2707 &	\bf 0.4915 &	\bf 0.6305 \\
 &\code{T5$^*$}   &  0.0137 &	0.0042 & 0.0200 &	0.0266	& 0.0237 &	0.3735 \\
  &  \code{BART$^*$}   & 0.0793 &	0.0053 &	0.0347 &	0.0872 &	0.0153 &	0.3181  \\
 &  \code{COMET$^*$}   & 0.0562 &	0.0216 &	0.0474 &	0.0902 &	0.0862 &	0.4402\\
 & \code{GLUCOSE-T5$^*$}   & 0.0654 &	0.0287 &	0.0560 &	0.0827 &	0.1332 &	0.4442 \\
 \midrule

\rotatetabularnormal{8}{ghostwhite}{(1.1.2)}{SE}& 
\code{T5} &  \textbf{0.3083}  & 	 \textbf{0.1619}  & 	 \textbf{0.1662}  & 	 0.2760  & 	0.4119  & 		 0.6276 \\
  &  \code{BART} &  0.2926  & 	 0.1484  & 	 0.1608  & 	 0.2670  & 	 0.3681    & 	 0.6166 \\
 &  \code{COMET} & 0.3053 &	0.1565 & 0.1588 &	0.2730 &	0.3850 &	0.6211 \\
 & \code{GLUCOSE-T5} & 0.3000 &	0.1611 &	0.1628 &	\textbf{0.2778} &	\textbf{0.4430} &	\textbf{0.6297} \\
 &\code{T5$^*$}   &  0.0133 &	0.0045 &	0.0191 &	0.0264 &	0.0241 &	0.3865 \\
  &  \code{BART$^*$}   & 0.0823 &	0.0061 & 0.0345 &	0.0926 &	0.0140 &	0.3243  \\
 &  \code{COMET$^*$}   & 0.0567 &	0.0217 &	0.0472 &	0.0937 &	0.0884 &	0.4523 \\
 & \code{GLUCOSE-T5$^*$}   & 0.0003 &	0.0001 &	0.0070 &	0.0024 &	0.0032 &	0.3073\\
 
 \midrule

\rotatetabularnormal{8}{ghostwhite}{(1.1.3)}{SE Clipped}&
 \code{T5} &  0.2889  & 	 0.1448  & 	 \textbf{0.1549}  & 	 0.2618  & 	 0.3099  & 	\textbf{0.6123} \\
  &  \code{BART} &  0.2651  & 	 0.1272  & 	 0.1384  & 	 0.2409  & 	 0.2765  &  	 0.5814 \\
 &  \code{COMET} & \textbf{0.3023}  &	\textbf{0.1509}   &	0.1536  &	\textbf{0.2667}  &	0.3090  &	0.6083 \\
 & \code{GLUCOSE-T5}  & 0.2870  &	0.1461  &	0.1523  &	0.2645  &	\textbf{0.3238}  &	0.6094\\
 &\code{T5$^*$}   & 0.0559  &	0.0199  &	0.0439  &	0.0564  &	0.0762  &	0.4549  \\
  &  \code{BART$^*$}   & 0.0931  &	0.0067  &	0.0367  &	0.0869  &	0.0198  &	0.3541  \\
 &  \code{COMET$^*$}   & 0.0577  &	0.0215  &	0.0479  &	0.0953  &	0.0911  &	0.4583 \\
 & \code{GLUCOSE-T5$^*$}    & 0.0003  &	0.0001  &	0.0066  &	0.0025  &	0.0034  &	0.3063\\

\midrule

\rotatetabularnormal{8}{Green2}{(1.2.1)}{Prerequisite}& 
 \code{T5} & 0.1826  & 	 0.1002  & 	 0.1282  & 	0.2176  & 	 0.3357  &  	\textbf{ 0.5902} \\
  &  \code{BART} & 0.1817  & 	 0.1020  & 	 0.1260  & 	 0.2118  & 	\textbf{0.3401}  & 	 0.5804 \\
 &  \code{COMET} & \textbf{0.2115} &	\textbf{0.1145} &	0.1296 &	0.2168 &	0.3064 &	0.5815 \\
 & \code{GLUCOSE-T5} & 0.1812 &	0.1001 &	\textbf{0.1299} &	\textbf{0.2197} &	0.3144 &	0.5896 \\
 &\code{T5$^*$}   &  0.0177 &	0.0043 &	0.0222 &	0.0279 &	0.0225 &	0.3541 \\
  &  \code{BART$^*$}   & 0.0779 &	0.0065 &	0.0334 &	0.0827 &	0.0166 &	0.2913  \\
 &  \code{COMET$^*$}   & 0.0517 & 	0.0186 &	0.0447 &	0.0782 &	0.0768 &	0.4281\\
 & \code{GLUCOSE-T5$^*$}    &0.0259 &	0.0108 &	0.0394 &	0.0625 &	0.0889 &	0.4392\\

\midrule

\rotatetabularnormal{8}{Blue2}{(1.2.2)}{Motivation}&
  \code{T5} & 0.3462  & 	 0.2503  & 	 0.1998  & 	 0.3781  & 	 0.7109  & 		 0.6973 \\
  &  \code{BART} &  0.3497  & 	 0.2482  & 	 0.1961  & 	 0.3709  & 	 0.6434  &	 0.6914  \\
 &  \code{COMET} & 0.3428 &	0.2381 &	0.1935 &	0.3649 &	0.6286 &	0.6962 \\
 & \code{GLUCOSE-T5} & \textbf{0.3546} &	\textbf{0.2582} &	\textbf{0.2037} &	\textbf{0.3840} &	\textbf{0.7499} &	\textbf{0.7048} \\
 &\code{T5$^*$}   &  0.0134 &	0.0033 & 0.0183 &	0.0257 &	0.0181 &	0.4038 \\
  &  \code{BART$^*$}   & 0.1072 &	0.0082 &	0.0416 &	0.1212 &	0.0164 &	0.3497  \\
 &  \code{COMET$^*$}   & 0.0582 &	0.0215 &	0.0475 &	0.0882 &	0.0782 &	0.4516 \\
 & \code{GLUCOSE-T5$^*$}    & 0.0504 &	0.0174 &	0.0434 &	0.0632 &	0.0696 &	0.4053\\
 
 \midrule

\rotatetabularnormal{8}{Gray3}{(1.2.3)}{Reaction} &
  \code{T5} &  \textbf{0.3410}  & 	 \textbf{0.2397}  &  	 \textbf{0.1939}  & 	 \textbf{0.3720}  & 	 0.5177  & 	 \textbf{0.6665}  \\
  &  \code{BART} &  0.3320  & 	 0.2297  & 	 0.1869  & 	 0.3531  & 	 0.4575  &  	 0.6575  \\
 &  \code{COMET} & 0.3338 &	0.2273 &	0.1815 &	0.3406 &	0.2662 &	0.6520\\
 & \code{GLUCOSE-T5} & 0.3283 &	0.2318 &	0.1903 &	0.3716 &	\textbf{0.5364} &	0.6653\\
 &\code{T5$^*$}   &  0.0116 &	0.0037 &	0.0201 &	0.0239 &	0.0167 &	0.3899 \\
  &  \code{BART$^*$}   & 0.1815 &	0.0418 &	0.0913 &	0.1531 &	0.0194 &	0.5353  \\
 &  \code{COMET$^*$}   & 0.0590 &	0.0204 &	0.0454 &	0.0966 &	0.0653 &	0.4299\\
 & \code{GLUCOSE-T5$^*$}    & 0.0534 &	0.0213 &	0.0459 &	0.0759 &	0.0719 &	0.4125\\
 
\cmidrule{1-8}
\rotatetabularnormal{8}{brilliantlavender}{Average}{Score}&
 \code{T5} & 0.2924 & 0.1744 & \textbf{0.1677} & 0.2947 & 0.4570 & 0.6369 \\
 & \code{BART} & 0.2792 & 0.1658 & 0.1602 &  0.2837 & 0.4183  & 0.6249 \\
 & \code{COMET} & \textbf{0.2953} & 0.1732 & 0.1625 &  0.2879 & 0.3906  &   0.6307 \\
 & \code{GLUCOSE-T5} & 0.2908 & \textbf{0.1756} & 0.1671 & \textbf{0.2980} & \textbf{0.4765} & \textbf{0.6382} \\
 & \code{T5$^*$} & 0.0209 & 0.0066 & 0.0239 & 0.0312 & 0.0302 & 0.3938 \\
  & \code{BART$^*$} & 0.1036 & 0.0124 & 0.0454 &  0.1040 & 0.0169 & 0.3621 \\
  & \code{COMET$^*$} & 0.0575 & 0.0185 & 0.0445 &  0.0917 & 0.0641 & 0.4303 \\
 & \code{GLUCOSE-T5$^*$} & 0.0310 & 0.0130 & 0.0322 & 0.0491 & 0.0601 & 0.3886 \\

\bottomrule
\end{tabular}
}
\caption{Results for Task 1. \code{T5$^*$}, \code{BART$^*$}, \code{COMET$^*$} and \code{GLUCOSE-T5$^*$} are not fine-tuned on \dataset{}. \colorbox{ghostwhite}{SE} denotes Subsequent Event.}
\label{tab:results-sup}
\end{table*}

\begin{table*}[t]
\centering
\resizebox{\linewidth}{!}{
\begin{tabular}{l|cccccc}
\toprule
\textbf{Model} & \textbf{BLEU1} & \textbf{BLEU2} & \textbf{METEOR} & \textbf{ROUGE} & \textbf{CIDEr} & \textbf{Sem-Sim} \\
\midrule
\colorbox{almond}{\textbf{(1.1.4) Chained Cause}} & & \\
\quad \quad \code{T5} & 0.2781  & 	 0.1566   & 	0.1675  & 	 0.2757  & 	 0.5303  & 	 0.6518 \\
  \quad \quad  \code{BART} &  0.1960  &	0.1104  &	0.1382  &	0.2242  &	0.4231  &	0.6074 \\
\quad \quad \code{COMET} &  \textbf{0.2893}  &	\textbf{0.1633}  &	0.1674  &	0.2742  &	0.5247  &	0.6488 \\
\quad \quad \code{GLUCOSE-T5}  & 0.2820  &	0.1600  &	\bf 0.1697  &	\bf 0.2796  &	\bf 0.5633  &	\bf 0.6557\\
\cmidrule{2-7}
\colorbox{almond}{\textbf{(1.1.1)* Cause}} & & \\
\quad \quad \code{T5} & 0.2884  & 	 0.1503   & 	 0.1635  & 	 0.2634  &  0.4591  & 0.6284  \\
  \quad \quad  \code{BART} &  0.2548  &	0.1400   &	0.1530  &	0.2590  &	0.4279  &	0.6225 \\
\quad \quad \code{COMET} &  0.2769  &	0.1522  &	0.1584  &	0.2654  &	0.4510  &	0.6257 \\
\quad \quad \code{GLUCOSE-T5}  & \bf 0.2938  &	\bf 0.1564  &	\bf 0.1636  &	\bf 0.2709  &	\bf 0.4915  &	\bf 0.6310\\
\midrule
\colorbox{ghostwhite}{\textbf{(1.1.5) Chained SE}} & & \\
\quad \quad \code{T5} & \textbf{0.3322}  & 	 \textbf{0.1813}   & 	 \textbf{0.1784}  & 	 0.2940  & 	 0.5136  & 	 0.6469 \\
  \quad \quad  \code{BART} &  0.3131  &	0.1649   &	0.1672  &	0.2795  &	0.4106  &	0.6314 \\
\quad \quad \code{COMET} &  0.3057  &	0.1626  &	0.1673  &	0.2742  &	0.4515  &	0.6321 \\
\quad \quad \code{GLUCOSE-T5}  & 0.3258  &	0.1789  &	0.1776  &	\bf 0.2943  &	\bf 0.5218  &	\bf 0.6516\\
\cmidrule{2-7}
\colorbox{ghostwhite}{\textbf{(1.1.2)* SE}} & & \\
\quad \quad \code{T5} & \bf 0.3088  & 	 \bf 0.1622  & 	 0.0841   & 	 0.2764  & 	 0.4167  & 0.6279  \\
  \quad \quad  \code{BART} & 0.2919  &	0.1490  & 0.1617  &	0.2667  &	0.3719  &	0.6165  \\
\quad \quad \code{COMET} &  0.3036  &	0.1557  &	0.1580  &	0.2727  &	0.3790&	0.6187 \\
\quad \quad \code{GLUCOSE-T5}  & 0.2998  &	0.1612  &	\bf 0.1628  &	\bf 0.2778  &	\bf 0.4471  &	\bf 0.6294\\
\bottomrule
\end{tabular}
}
\caption{Results for chained cause effect generation. (1.1.1)* and (1.1.2)* indicates results from Task 1.1.1, and 1.1.2 (as in \cref{tab:results-sup}), but only for target instances which have both cause and effect annotated, ensuring a fair comparison with (1.2). \colorbox{ghostwhite}{SE} denotes Subsequent Event.}
\label{tab:cqa-sup}
\end{table*}

\begin{table}[h!]
\centering
\resizebox{\linewidth}{!}{
\begin{tabular}{l|ccc}
\toprule
\textbf{Model}  & \textbf{Creativity} & \textbf{Contextuality}  & \textbf{Fluency} \\
\midrule
Gold & 4.7 & 4.8 & 5.0 \\
\midrule
\code{T5} & 3.8 & 4.1 & \bf 4.9 \\
\code{BART} & 3.6 & \bf 4.3 & \bf 4.9 \\
\code{COMET} & 3.8 & 4.1 & 4.8 \\
\code{GLUCOSE-T5} & \bf 3.9 & \bf 4.3 & \bf 4.9 \\
\code{T5}$^*$  & 2.4 & 2.1 & 1.9 \\
\code{BART}$^*$  & 2.6 & 2.5 & 1.8 \\
\code{COMET}$^*$  & 2.2 & 2.3 & 2.5 \\
\code{GLUCOSE-T5}$^*$  & 1.9 & 2.1 & 2.9 \\
\bottomrule
\end{tabular}%
}
\caption{Results of the human evaluation for the \dataset{}$_{NLG}$ task. \code{T5$^*$}, \code{BART$^*$}, \code{COMET$^*$}, and \code{GLUCOSE-T5$^*$} represent non fine-tuned versions.}
\label{tab:human-eval-sup}
\end{table}

\begin{table*}[ht!]
  \centering
  \begin{subtable}{\textwidth}
  \centering
  \resizebox{\textwidth}{!}{
    \begin{tabular}{p{16cm}}
    \toprule
    \textbf{A \pmb {($u_1$)}}: Hi, Jenny. Is it true you're moving to London?
    \textbf{B \pmb {($u_2$)}}: Yes, it is.
    \textbf{A \pmb {($u_3$)}}: What made you decide to do that?
    \textbf{B \pmb {($u_4$)}}: Work, mainly. I'm sure I'll be able to find a job there.
    \textbf{A \pmb {($u_5$)}}: You're probably right. But where are you going to live?
    \textbf{B \pmb {($u_6$)}}: I hope I'll find a flat to share with somebody. That way it will be cheaper.
    \textbf{A \pmb {($u_7$)}}: Yes, that's a good idea. Are you taking your dog with you?
    \textbf{B \pmb {($u_8$)}}:  No, I don't think so. My parents have offered to take care of him, and I don't think he'd be happy in the city.
    \textbf{A \pmb {($u_9$)}}: You're probably right. But aren't you afraid of moving to such a big place, especially after living in a small village?
    \textbf{B \pmb {($u_{10}$)}}: Not really. I think I'll enjoy myself. There's so much to do there; I expect I won't miss the countryside much and I can always come back and visit. 
    \textbf{A \pmb {($u_{11}$)}}: Well, I just hope you'll invite me to stay when you get settled.
    \textbf{B \pmb {($u_{12}$)}}:  Of course I will.
    \\
    \end{tabular}
    }
\end{subtable}
\bigskip
\begin{subtable}{\textwidth}
\centering
\resizebox{\textwidth}{!}{
  \begin{tabular}{p{16cm}}
      \toprule
      \textbf{Target -} \pmb{$u_6$}; \textbf{Inference:} \colorbox{almond}{\bf Cause}; \textbf{Answers:} \colorbox{gold}{\textbf{Gold}} - Being an expensive city, it is quite difficult to find an affordable place to live in London. \colorbox{Blue4}{\textbf{T5}} - The listener asked Jenny where she was going to live. \colorbox{brilliantlavender}{\textbf{COMET}} - The speaker is looking for a flat to live in London. \colorbox{aqua}{\textbf{GLUCOSE-T5}-} Jenny has decided to move to London for her job. \\
      
      \midrule
      \textbf{Target -} \pmb{$u_{10}$}; \textbf{Inference:} \colorbox{almond}{\bf Cause}; \textbf{Answers:} \colorbox{gold}{\textbf{Gold}}  - Jenny realizes that a city like London will provide a great quality of life for her. \colorbox{Blue4}{\textbf{T5}} - The listener asked Jenny if she was afraid of moving to London after living in a small village. \colorbox{brilliantlavender}{\textbf{COMET}} - The speaker is moving to London for a job.
      \colorbox{aqua}{\textbf{GLUCOSE-T5}-} The listener asked Jenny if she was afraid of moving to such a big place. \\
      
      \midrule
      \textbf{Target -} \pmb{$u_6$}; \textbf{Inference:} \colorbox{ghostwhite}{\bf Subsequent Event}; \textbf{Answers:} \colorbox{gold}{\textbf{Gold}} - The listener suggests Jenny to find potential flats or flatmates online. \colorbox{Blue4}{\textbf{T5}} - The speaker will find a flat to share with a friend. \colorbox{brilliantlavender}{\textbf{COMET}} - The speaker informed the listener that she will share the flat with someone else. \colorbox{aqua}{\textbf{GLUCOSE-T5}-} Jenny will find a flat to share with her friend. \\
      
      \midrule
      \textbf{Target -} \pmb{$u_{10}$}; \textbf{Inference:} \colorbox{ghostwhite}{\bf Subsequent Event}; \textbf{Answers:} \colorbox{gold}{\textbf{Gold}} - Jenny inquired a social club in London and ask for their membership to utilize her free time. \colorbox{Blue4}{\textbf{T5}} - The speaker told the listener that he would love to visit London. \colorbox{brilliantlavender}{\textbf{COMET}} - The speaker informed the listener that he will miss the countryside very much. \colorbox{aqua}{\textbf{GLUCOSE-T5}-} The speaker informed the listener that he would love to come back to London. \\
      
      \midrule
      \textbf{Target -} \pmb{$u_4$}; \textbf{Inference:} \colorbox{Green2}{\bf Prerequisite}; \textbf{Answers:} \colorbox{gold}{\textbf{Gold}} - Jenny has completed her studies. \colorbox{Blue4}{\textbf{T5}} - The speaker has a job in London. \colorbox{brilliantlavender}{\textbf{COMET}} - Jenny has applied for a job in London. \colorbox{aqua}{\textbf{GLUCOSE-T5}-} The speaker has a job in London. \\
      \midrule
      \textbf{Target -} \pmb{$u_{12}$}; \textbf{Inference:} \colorbox{Green2}{\bf Prerequisite}; \textbf{Answers:} \colorbox{gold}{\textbf{Gold}} - Jenny and the listener are good friends. \colorbox{Blue4}{\textbf{T5}} - Jenny has invited her friend to stay with her in London. \colorbox{brilliantlavender}{\textbf{COMET}} - Jenny has a place to stay in London. \colorbox{aqua}{\textbf{GLUCOSE-T5}-}The listener invited Jenny to stay in London. \\
      
      \midrule
      \textbf{Target -} \pmb{$u_6$}; \textbf{Inference:} \colorbox{Blue2}{\bf Motivation}; \textbf{Answers:} \colorbox{gold}{\textbf{Gold}} - Jenny is optimistic about having someone as her flatmate to save on rent. \colorbox{Blue4}{\textbf{T5}} - Jenny is hopeful of finding a flat to share with someone. \colorbox{brilliantlavender}{\textbf{COMET}} -  Jenny is optimistic about having someone as her flatmate. \colorbox{aqua}{\textbf{GLUCOSE-T5}-} Jenny is hopeful that she will find a flat to share with somebody. \\
      
      \midrule
      \textbf{Target -} \pmb{$u_{12}$}; \textbf{Inference:} \colorbox{Gray3}{\bf Reaction}; \textbf{Answers:} \colorbox{gold}{\textbf{Gold}} - The listener is happy for Jenny and looks forward to being invited to London by Jenny. \colorbox{Blue4}{\textbf{T5}} - The listener is happy for Jenny. \colorbox{brilliantlavender}{\textbf{COMET}} - The listener is happy to know that the speaker is moving to London. \colorbox{aqua}{\textbf{GLUCOSE-T5}-} The listener is excited to meet Jenny in London. \\
      \bottomrule
  \end{tabular}
  }
\end{subtable}
\caption{Inferences extracted from a sample dialogue.}
\label{tab:examples-sup}
\end{table*}
\begin{table*}[t]
\small
\centering
\resizebox{0.96\linewidth}{!}{
\begin{tabular}{l|c|c|cccccc}
\toprule
\bf Model & \bf \begin{tabular}[c]{@{}c@{}}Trained\\ On\end{tabular} & \bf \begin{tabular}[c]{@{}c@{}}Evaluated\\ On\end{tabular}  & \bf Cause & \bf SE & \bf Prereq. & \bf Motiv. & \bf Emo. Reac. & \bf Avg. \\
\midrule
\code{RoBERTa} & Single & Single & 83.34 & 83.17 & 79.48 & 86.33 & 84.26 & 83.28  \\
\code{ELECTRA} & Single & Single &  \bf 87.09  & \bf 86.09 &  \bf 85.15 & \bf 90.31 & \bf 86.11 & \bf 86.82 \\
\midrule
\code{T5} & Single & Single & 95.19 & \bf 95.29 & 94.93 & \bf 96.52 & 96.99 & 95.54 \\
\code{Unified QA} & Single  & Single  & \bf 95.85 & 94.99 & \bf 95.55  & 96.35 & \bf 97.22 & \bf 95.70\\
\midrule
\code{T5} & Multiple & Multiple & 20.04  & 20.45 & 15.94 & 25.26 & 26.72 & 20.62 \\
\code{Unified QA} & Multiple  & Multiple  & \bf 25.68 & \bf 21.64 & \bf 21.51 & \bf 30.93 & \bf 31.03 & \bf 24.33 \\
\midrule
\code{T5} & Single \& Multiple & Single \& Multiple &  \bf 78.18 & 74.72 & \bf 75.50 & \bf 82.51 & \bf 84.59 & \bf 77.68 \\
\code{Unified QA} & Single \& Multiple  & Single \& Multiple  & 78.12 & \bf 74.79 & 75.36 & 81.58 & 84.08 & 77.51\\
\midrule
\code{T5} & Single \& Multiple & Single &  \bf 93.20 & \bf 91.28 & \bf 91.27 & \bf 95.19 & \bf 95.14 & \bf 92.71 \\
\code{Unified QA} & Single \& Multiple  & Single  & 93.12 & 91.16 & 91.00 & 94.28 & 94.79 & 92.45 \\
\midrule
\code{T5} & Single \& Multiple & Multiple & \bf 3.50 & 2.77 & 3.59 & \bf 3.61 & \bf 6.03 & 3.38 \\
\code{Unified QA} & Single \& Multiple  & Multiple  & \bf \bf 3.50 & \bf 3.69 & \bf 3.98 & 2.58 & 4.31 & \bf 3.60 \\
\bottomrule
\end{tabular}
}
\caption{Results of the \dataset{}$_{MCQ}$ task. SE denotes subsequent event. Single $\xrightarrow{}$ Instances with single answer. Multiple $\xrightarrow{}$ Instances with multiple answers.}
\label{tab:appendix-alt}
\end{table*}

\begin{table*}[t]
\small
\centering
\resizebox{0.96\textwidth}{!}{
\begin{tabular}{l|c|c|cccccc}
\toprule
\bf Model & \bf \begin{tabular}[c]{@{}c@{}}Trained\\ On\end{tabular} & \bf \begin{tabular}[c]{@{}c@{}}Evaluated\\ On\end{tabular}  & \bf Cause & \bf SE & \bf Prereq. & \bf Motiv. & \bf Emo. Reac. & \bf Avg. \\
\midrule
\code{RoBERTa} & Single & Single & - & 78.31 & - & 80.94 & - & 79.02  \\
\code{ELECTRA} & Single & Single & - & \bf 82.02 & - & \bf 87.41 & - & \bf 83.46 \\
\midrule
\code{T5} & Single & Single & - & 94.23 & - & 95.61 & - & 94.60 \\
\code{Unified QA} & Single  & Single  &  - & \bf 94.38 &  - &  \bf 96.19 & - & \bf 94.87 \\
\midrule
\code{T5} & Multiple & Multiple & - & 16.49 & - & 24.23 & - & 18.07 \\
\code{Unified QA} & Multiple  & Multiple  & - & \bf 19.79 & - & \bf 24.74 & - & \bf 20.80\\
\midrule
\code{T5} & Single \& Multiple & Single \& Multiple & - & \bf 74.99 & - & 80.73 & -  & \bf 76.46 \\
\code{Unified QA} & Single \& Multiple  & Single \& Multiple  & - & 74.67 & - & \bf 80.80 & - & 76.24 \\
\midrule
\code{T5} & Single \& Multiple & Single & - & \bf 91.95 & - & 93.29 & -  & \bf 92.31 \\
\code{Unified QA} & Single \& Multiple  & Single  & - & 91.43 & - & \bf 93.37 & - & 91.95 \\
\midrule
\code{T5} & Single \& Multiple & Multiple & - & 1.32 & - & \bf 2.58 & -  & 1.58 \\
\code{Unified QA} & Single \& Multiple  & Multiple  & - & \bf 1.85 & - & \bf 2.58 & - & \bf 2.00 \\
\bottomrule
\end{tabular}
}
\caption{Results of the \dataset{}$_{MCQ}$ task under the zero-shot setting. SE denotes subsequent event. Instance corresponding to cause, prerequisite, and emotional reaction are used for training. Instance corresponding to subsequent event and motivation are used for evaluation. Single $\xrightarrow{}$ Instances with single answer. Multiple $\xrightarrow{}$ Instances with multiple answers.}
\label{tab:appendix-alt2}
\end{table*}

\begin{figure}[t]
    \centering
        \includegraphics[width=\linewidth]{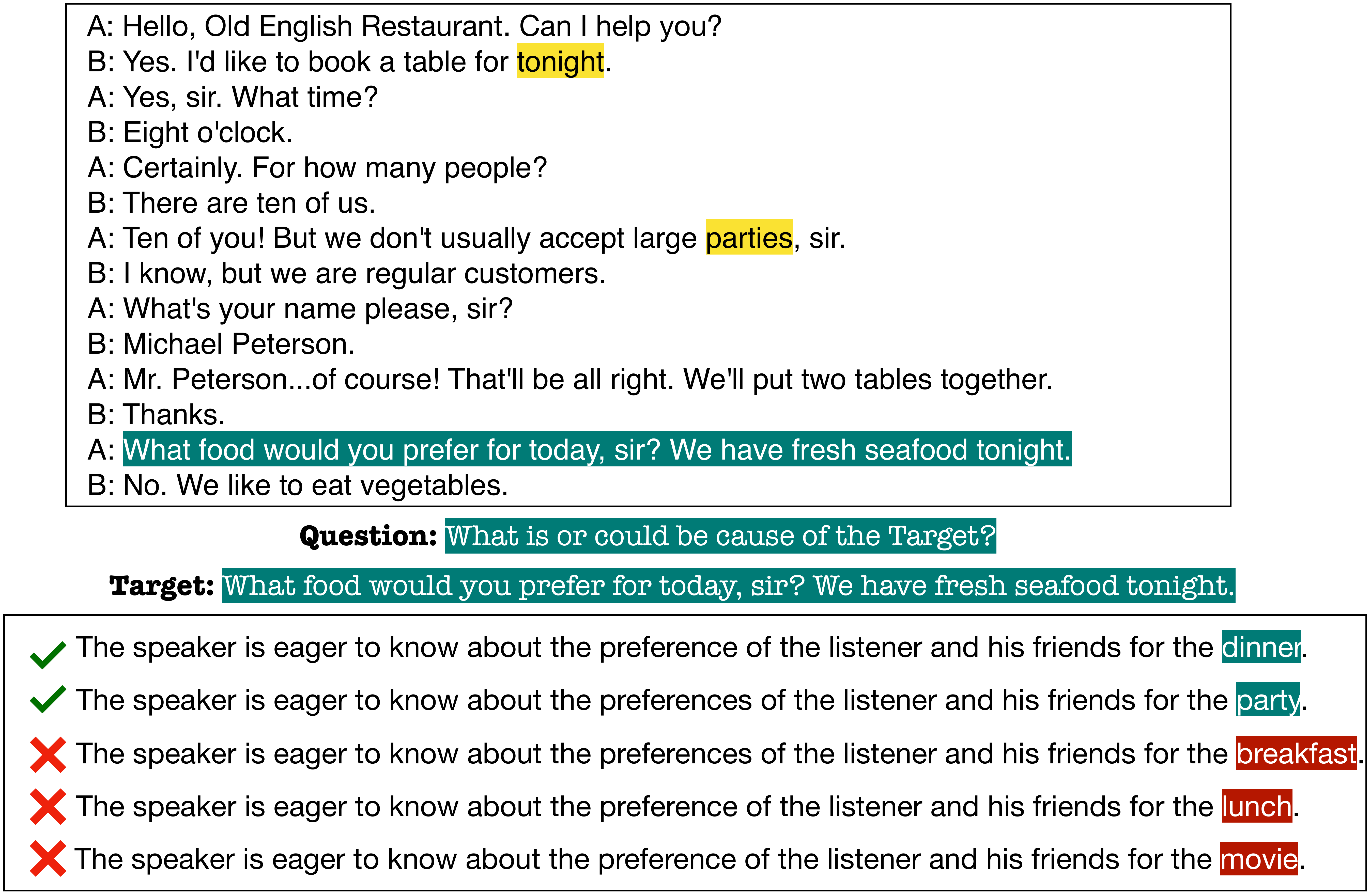}
        \caption{A data sample of \dataset{} for the \dataset{}$_{MCQ}$ task. Here, commonsense is required to infer the following events -- booking a table at night implies the intention of having dinner.}
        \label{fig:mcq2}
\end{figure}

\begin{figure}[!ht]
\begin{subfigure}[t]{\linewidth}
    \centering
        \includegraphics[width=\linewidth]{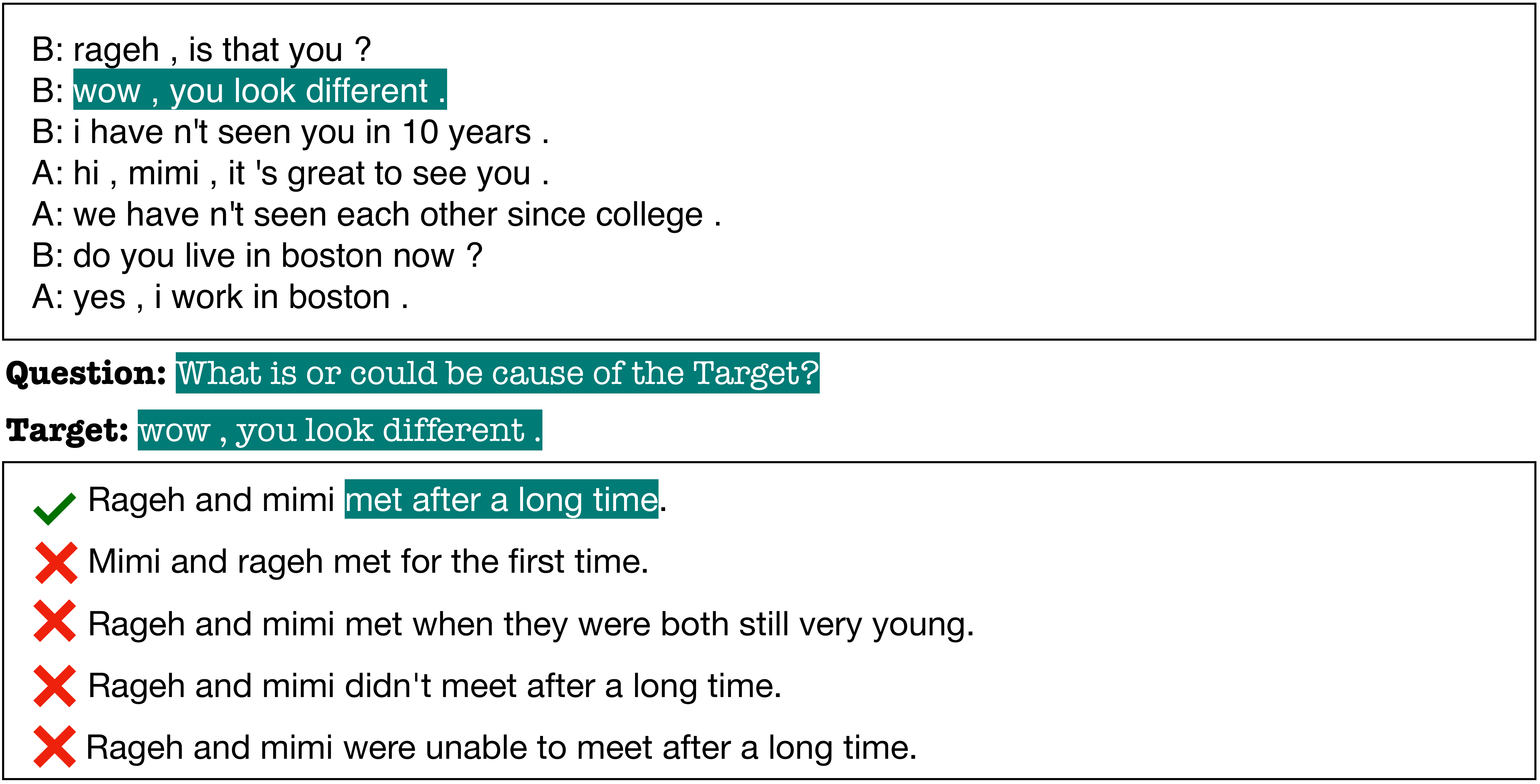}
        \caption{}
        \label{fig:mcq15}
\end{subfigure}
\begin{subfigure}[t]{\linewidth}
    \centering
        \includegraphics[width=\linewidth]{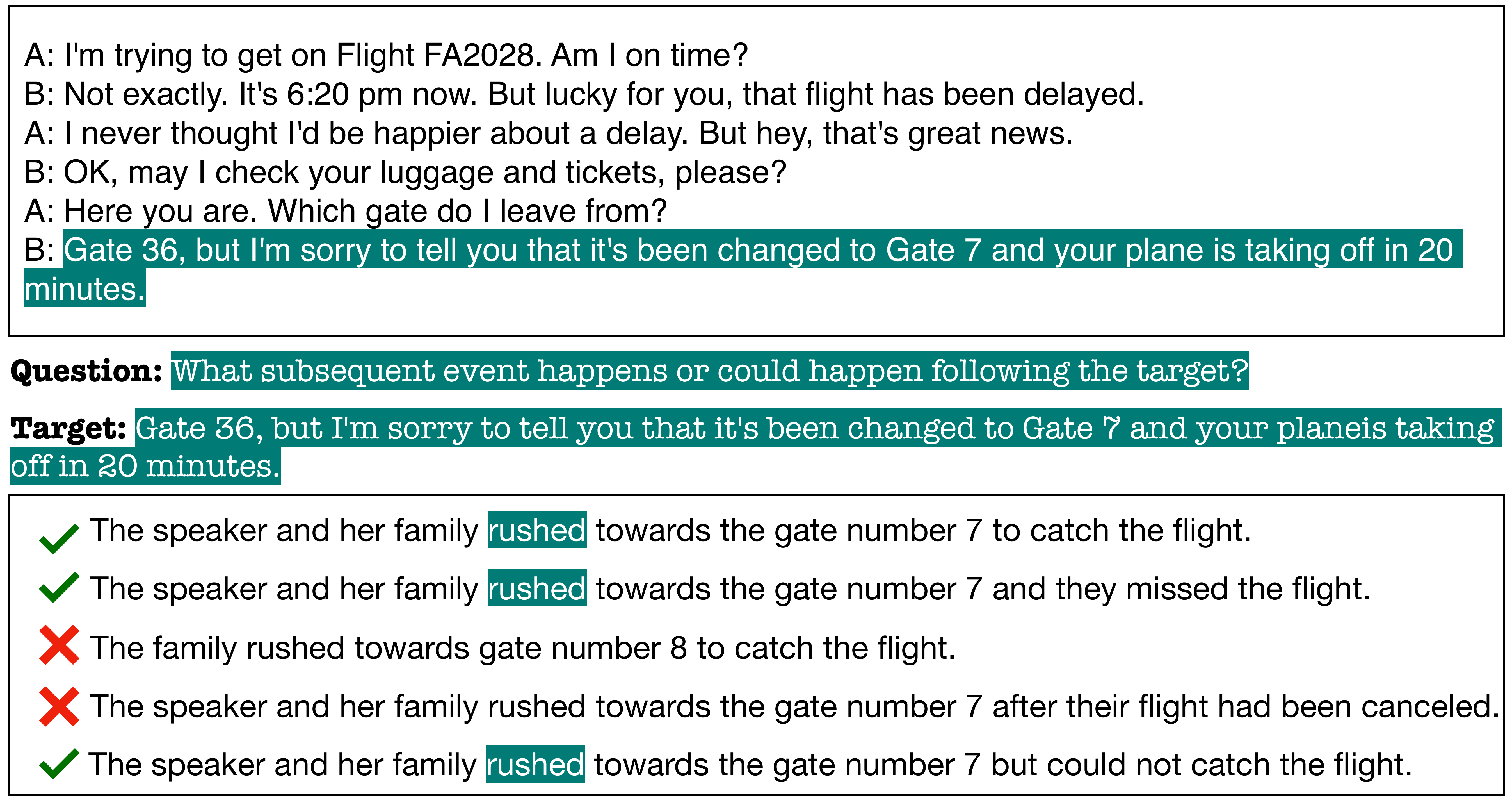}
        \caption{}
        \label{fig:mcq16}
\end{subfigure}
\caption{Instances of \texttt{temporal commonsense} in \dataset{}.}
\label{fig:temporal-csk-sup}
\end{figure}

\begin{figure}[t]
    \centering
        \includegraphics[width=\linewidth]{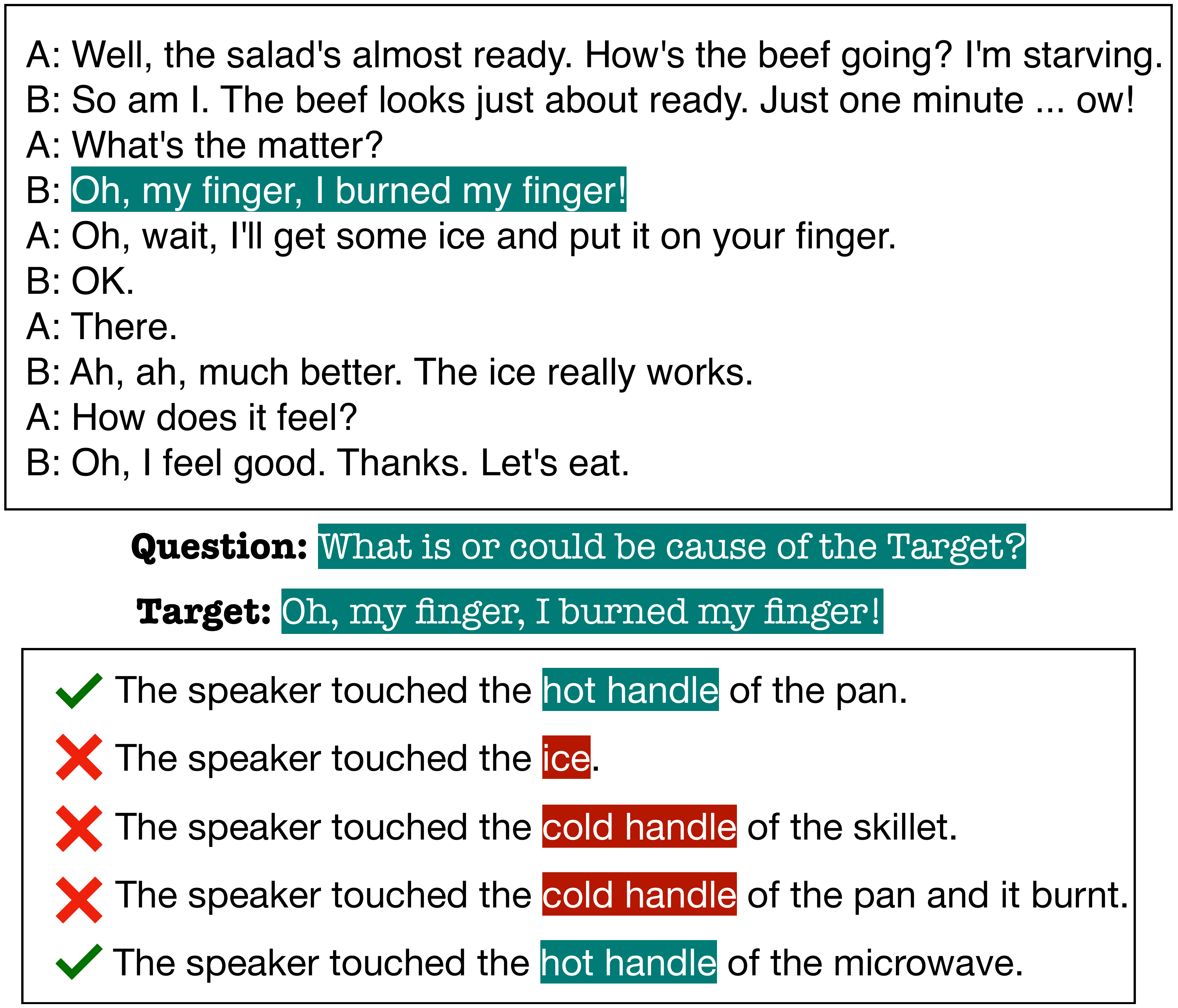}
        \caption{A data sample of \dataset{} where physical commonsense inference is prevalent.}
        \label{fig:mcq3}
\end{figure}

\begin{figure*}[ht]
    \centering
    
\begin{subfigure}[t]{.49\textwidth}
        \includegraphics[width=\linewidth]{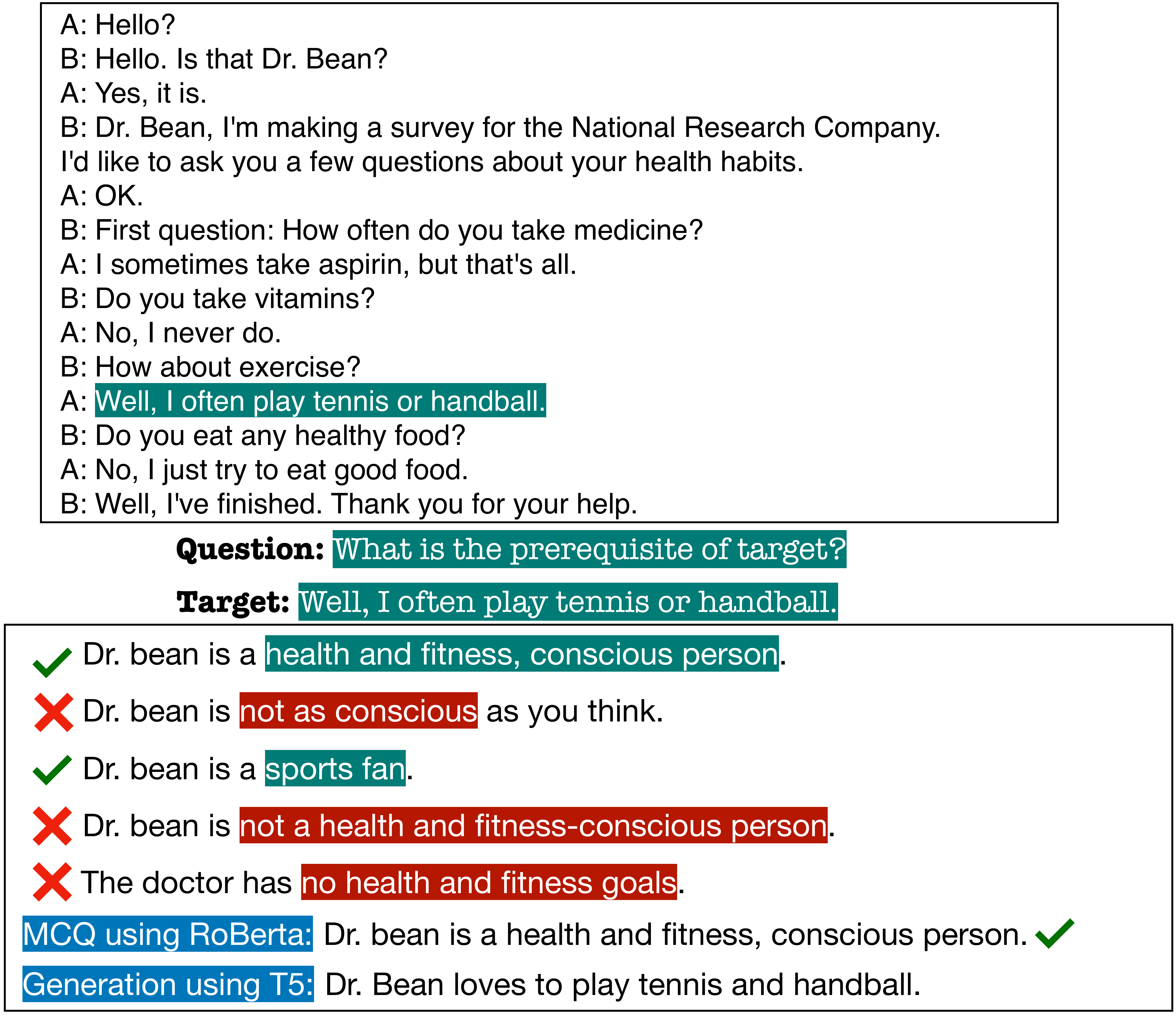}
        \caption{}
        \label{fig:mcq5}
\end{subfigure}~
\begin{subfigure}[t]{.49\textwidth}
        \includegraphics[width=\linewidth]{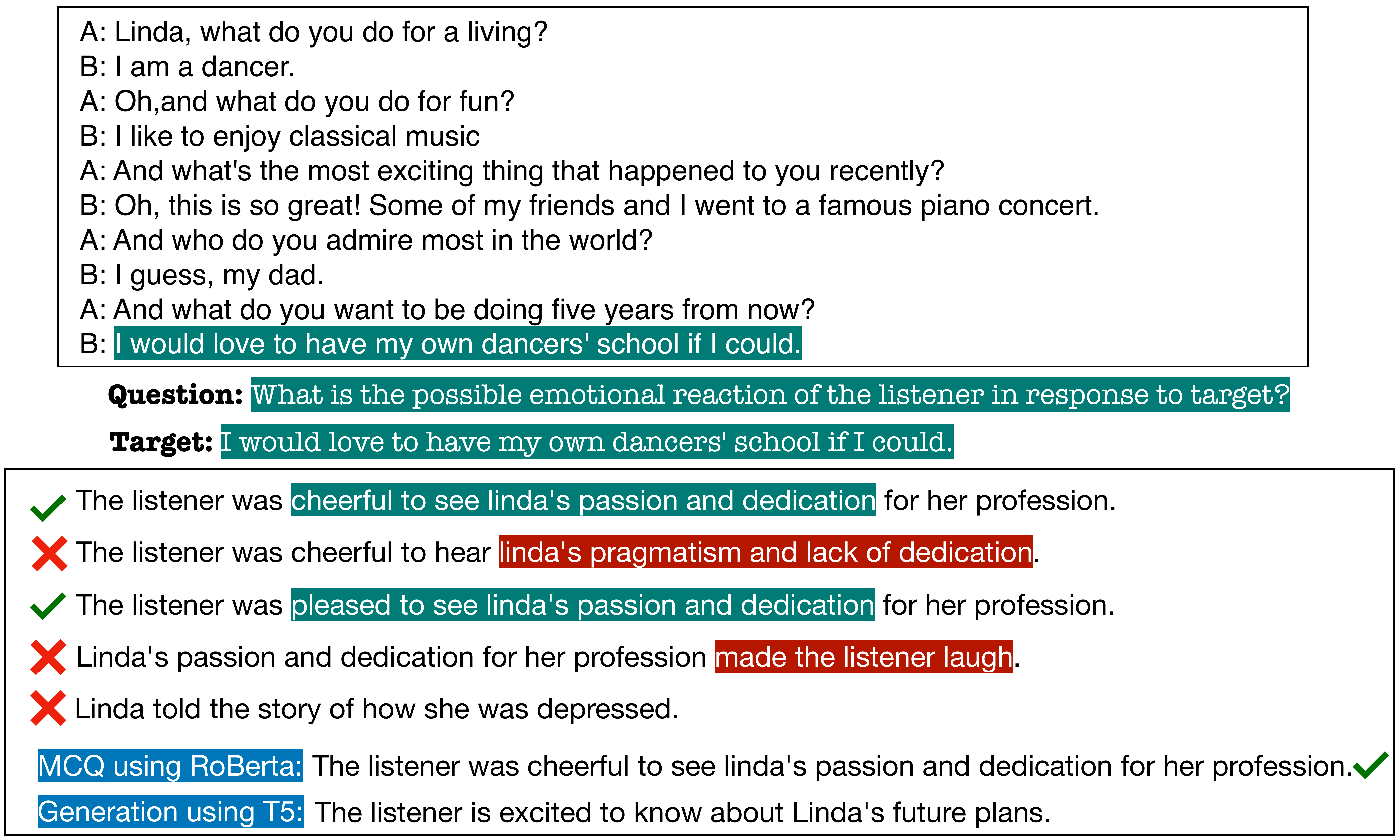}
        \caption{}
        \label{fig:mcq6}
\end{subfigure}

\begin{subfigure}[t]{.49\textwidth}
        \includegraphics[width=\linewidth]{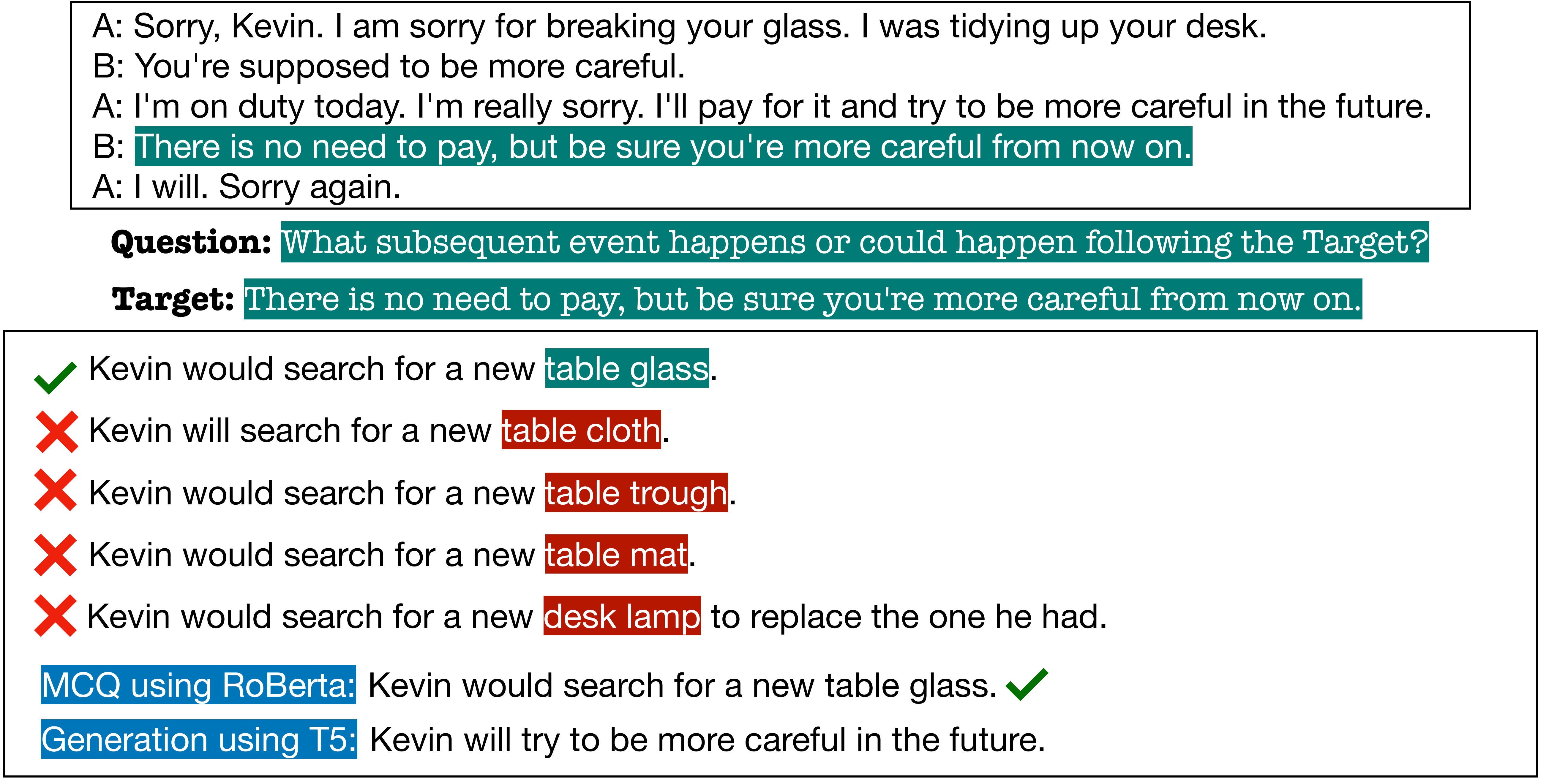}
        \caption{}
        \label{fig:mcq8}
\end{subfigure}~
\begin{subfigure}[t]{.49\textwidth}
        \includegraphics[width=\linewidth]{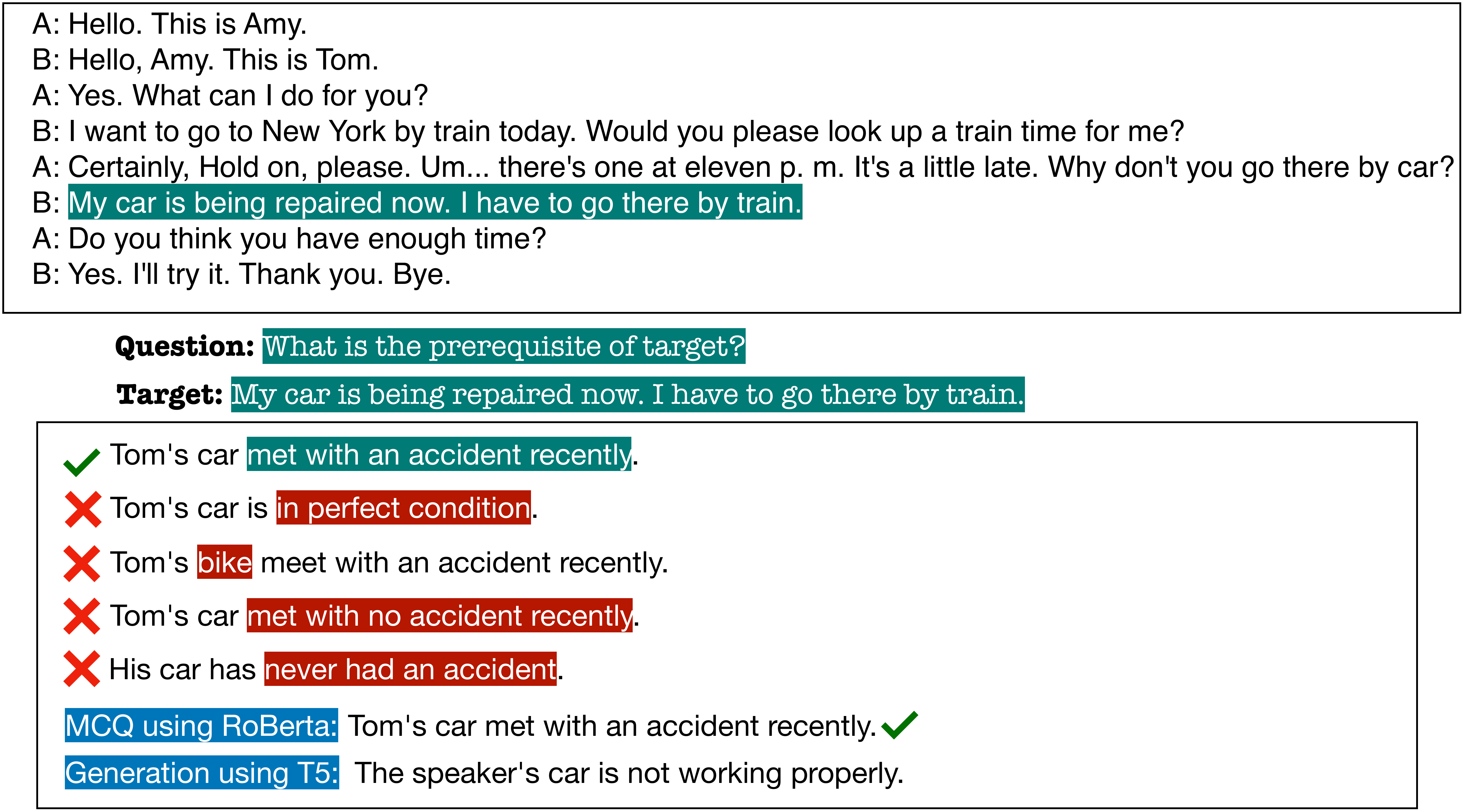}
        \caption{}
        \label{fig:mcq9}
\end{subfigure}
\caption{Instances of \texttt{general commonsense} in \dataset{}.}
\label{fig:general_csk_sup}
\end{figure*}

\begin{figure}[ht]
        \includegraphics[width=\linewidth]{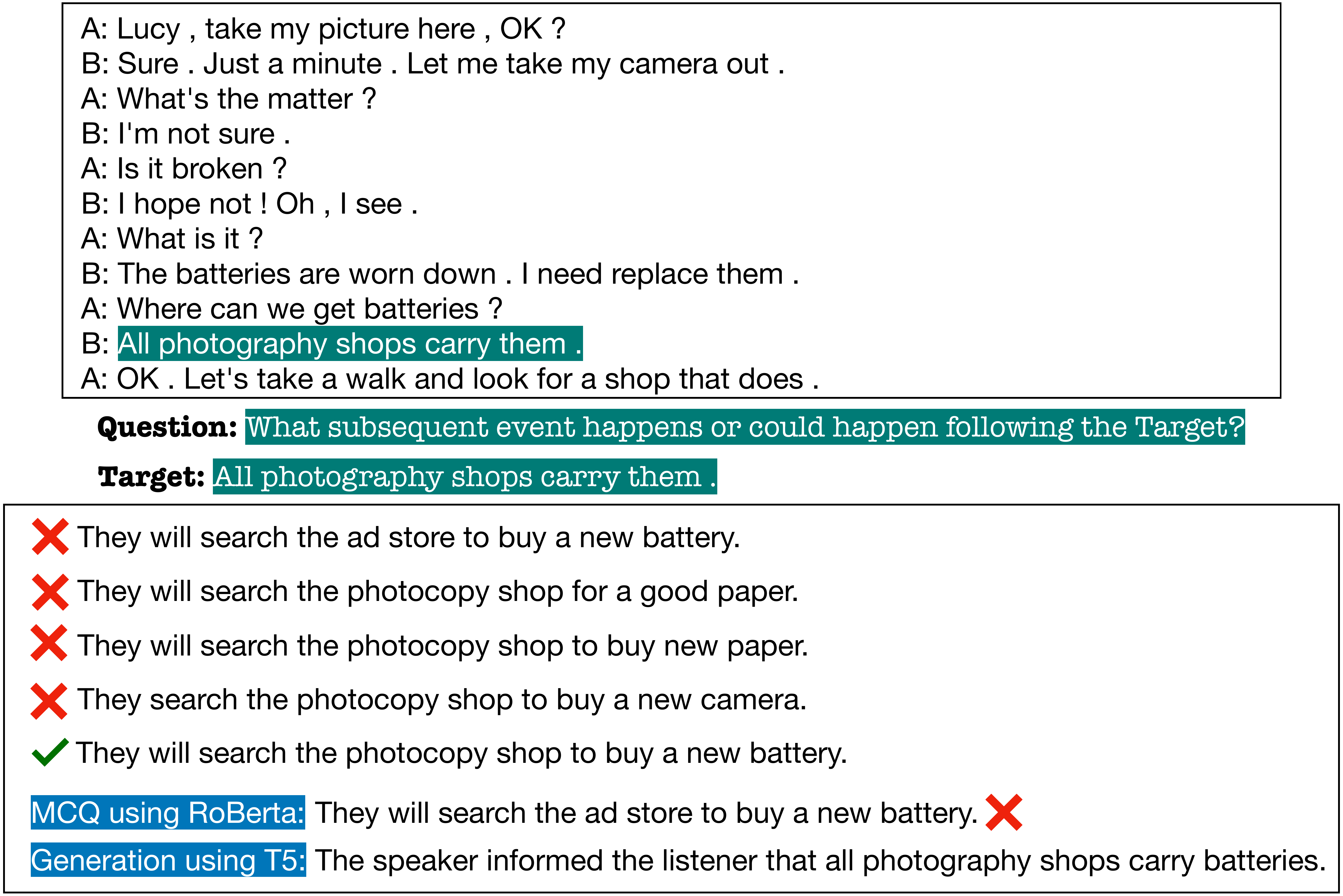}
        \caption{An instance where \code{RoBERTa} fails to capture the contextual commonsense cue.}
        \label{fig:mcq10}
\end{figure}
\begin{figure}[t]
\begin{subfigure}[t]{\linewidth}
    \centering
        \includegraphics[width=\linewidth]{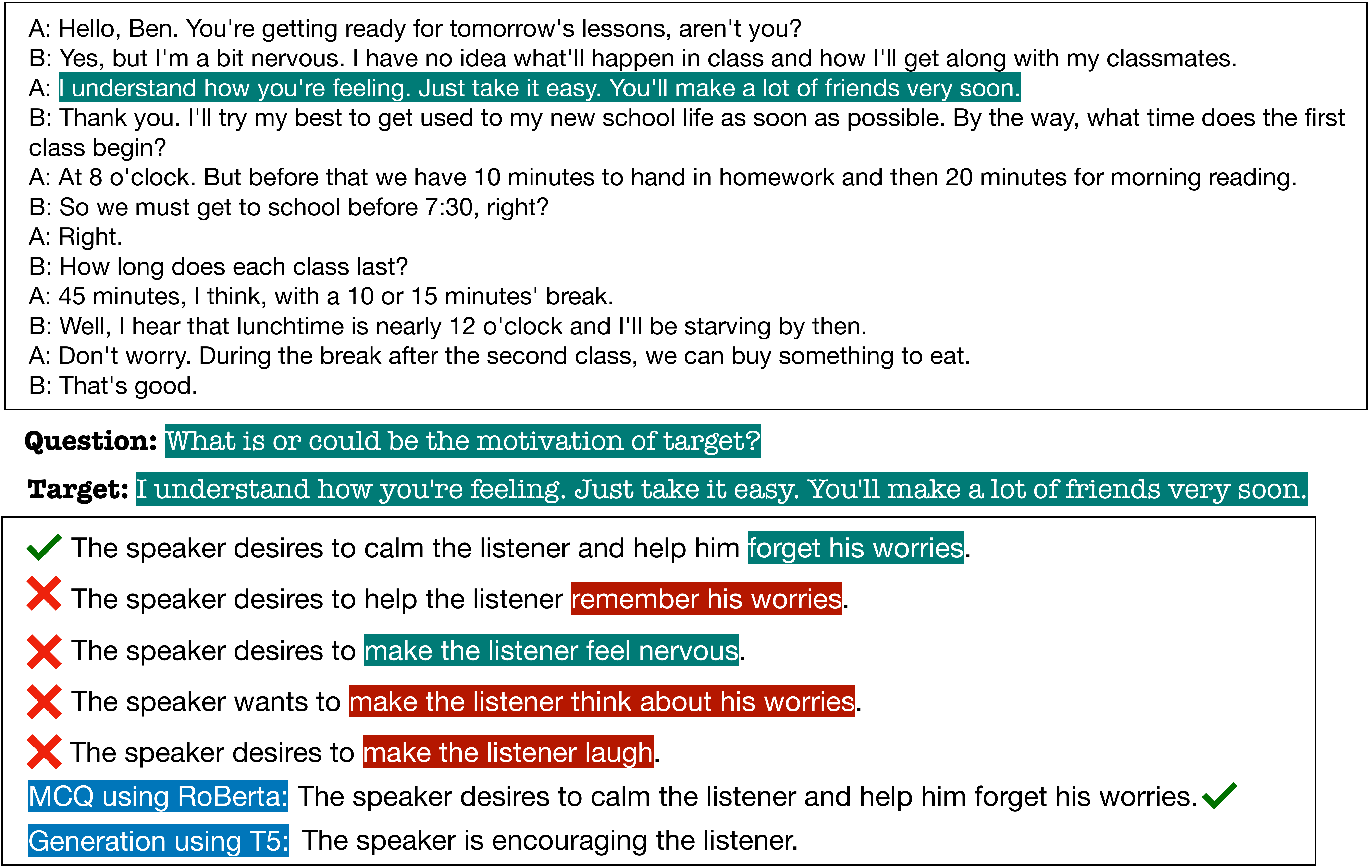}
        \caption{}
        \label{fig:mcq4}
\end{subfigure}

\begin{subfigure}[t]{\linewidth}
    \centering
        \includegraphics[width=\linewidth]{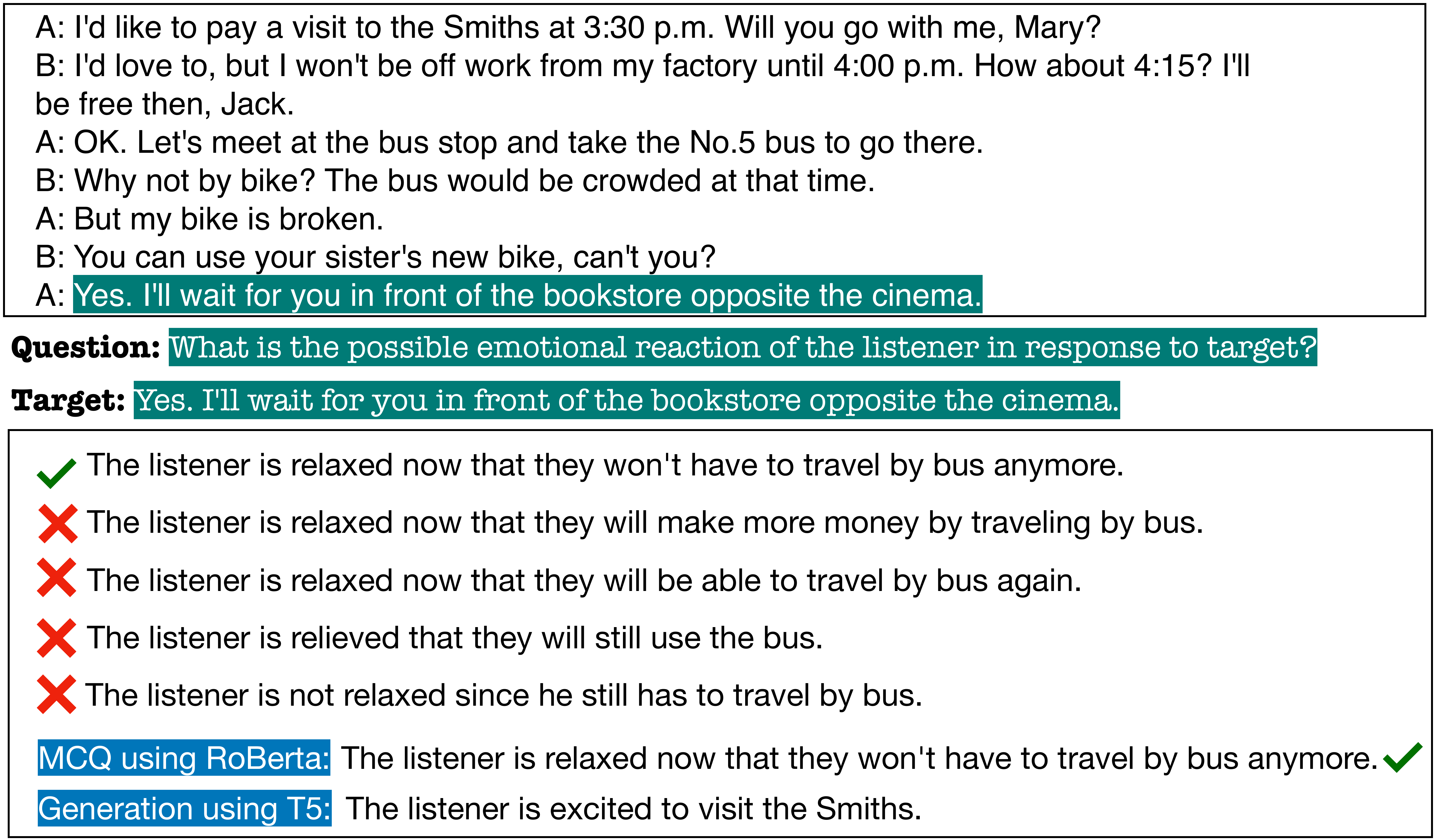}
        \caption{}
        \label{fig:mcq7}
\end{subfigure}
\caption{Instances of \texttt{social commonsense} in \dataset{}.}
\label{fig:social_csk_sup}
\end{figure}

\begin{figure}[ht]
\begin{subfigure}[t]{\linewidth}
    \centering
        \includegraphics[width=\linewidth]{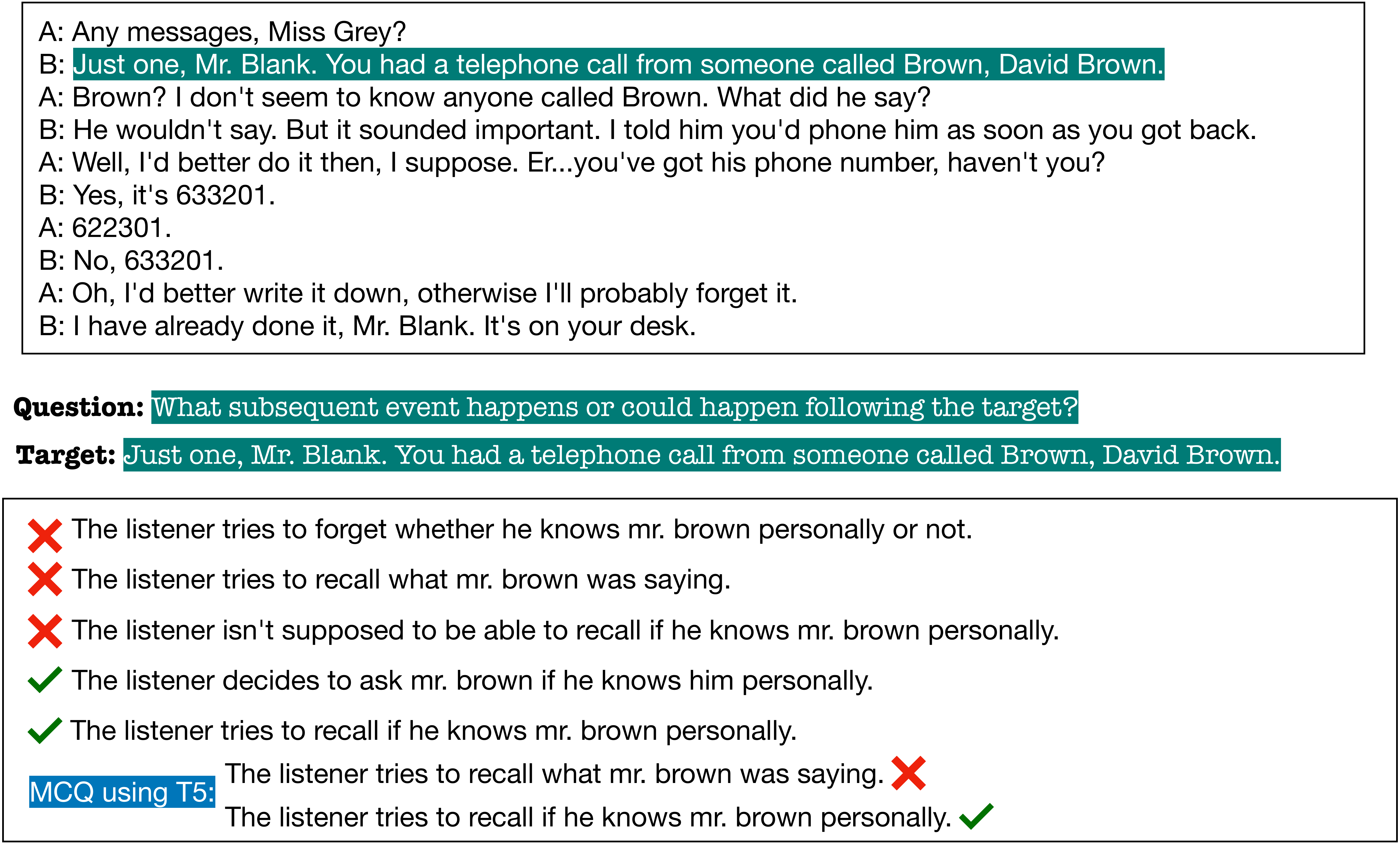}
        \caption{}
        \label{fig:mcq13}
\end{subfigure}
\begin{subfigure}[t]{\linewidth}
    \centering
        \includegraphics[width=\linewidth]{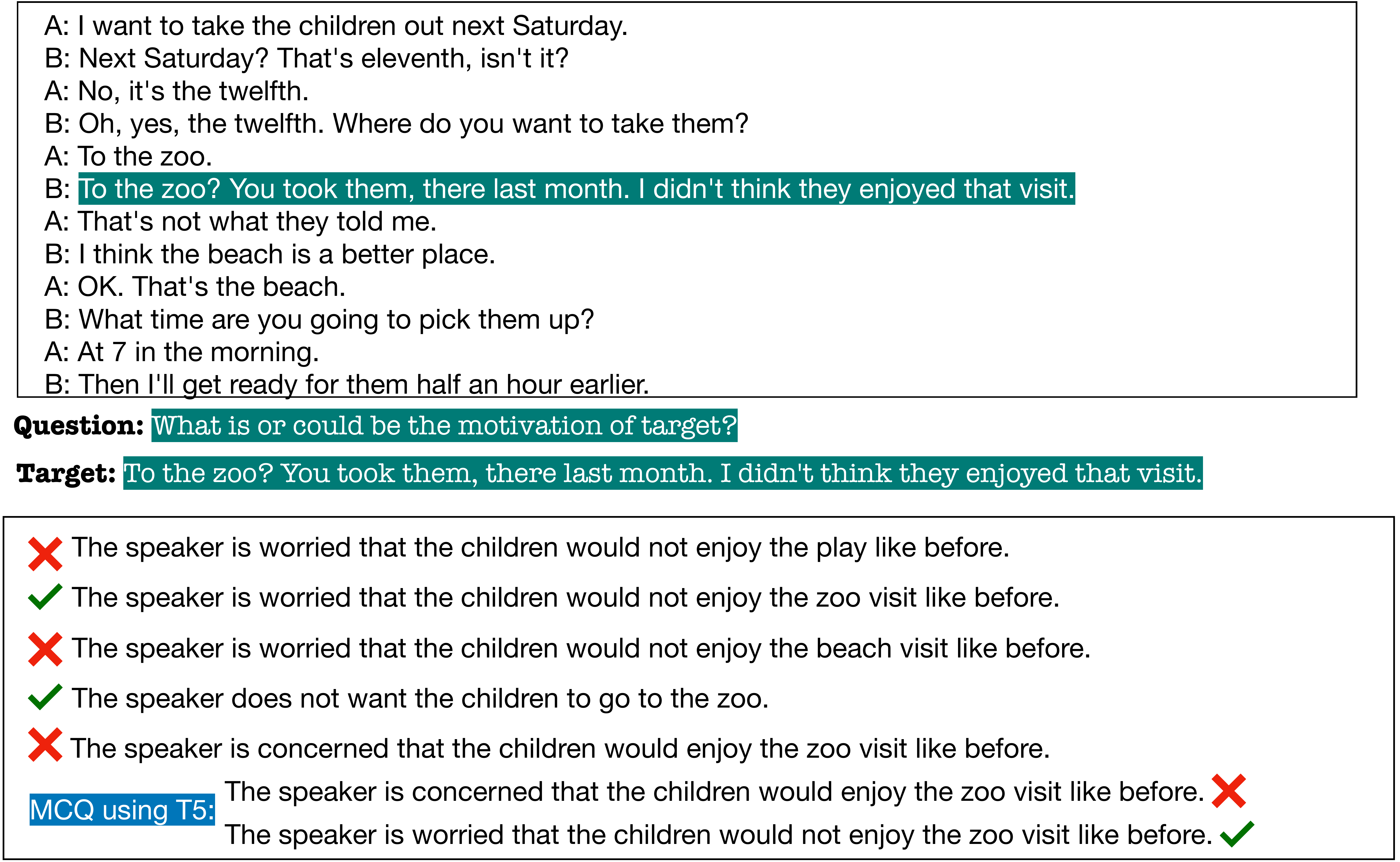}
        \caption{}
        \label{fig:mcq14}
\end{subfigure}
\caption{Multiple-answer predictions by \code{T5} for the \dataset{}$_{MCQ}$ task.}
\end{figure}

\begin{figure}[ht]
\begin{subfigure}[t]{\linewidth}
    \centering
        \includegraphics[width=\linewidth]{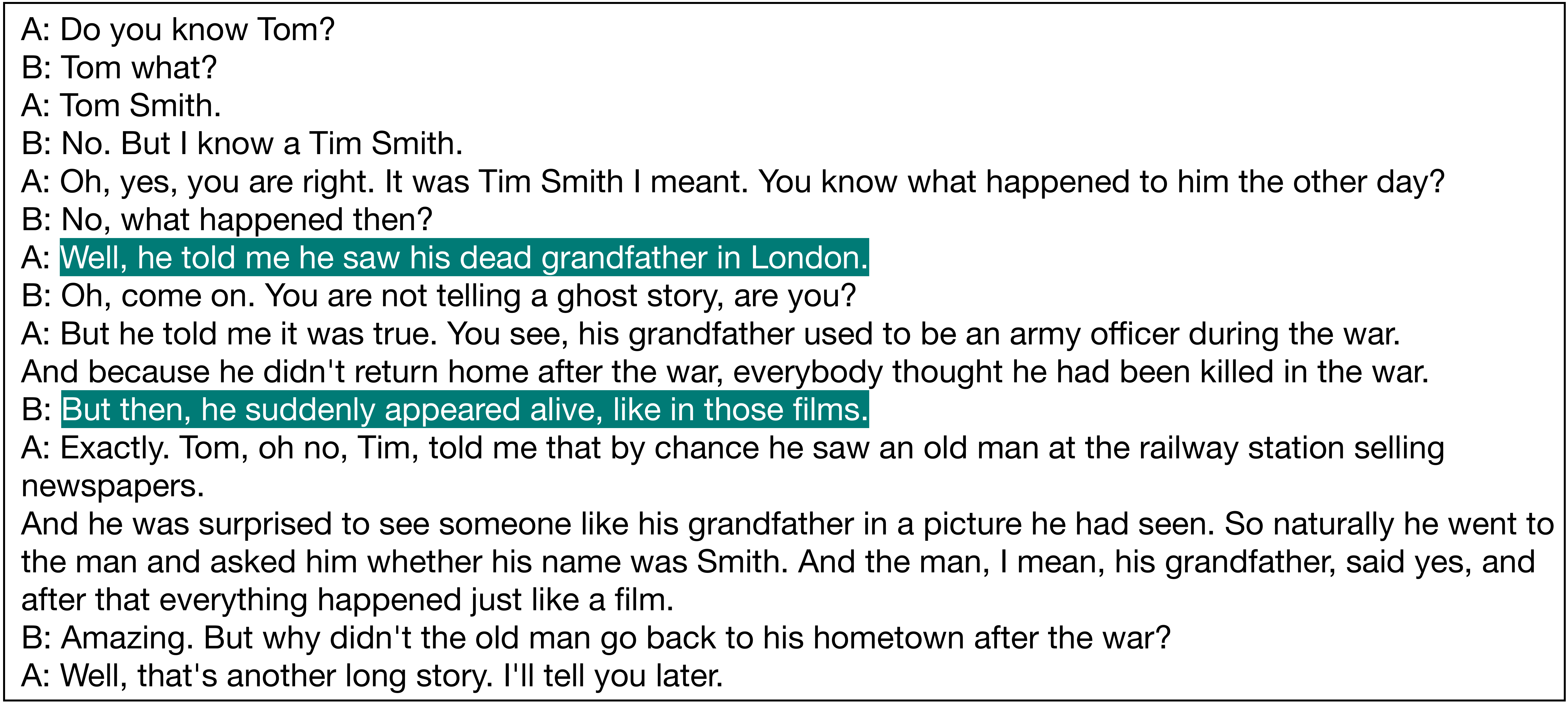}
        \label{fig:mcq11}
\end{subfigure}
\begin{subfigure}[t]{\linewidth}
    \centering
        \includegraphics[width=\linewidth]{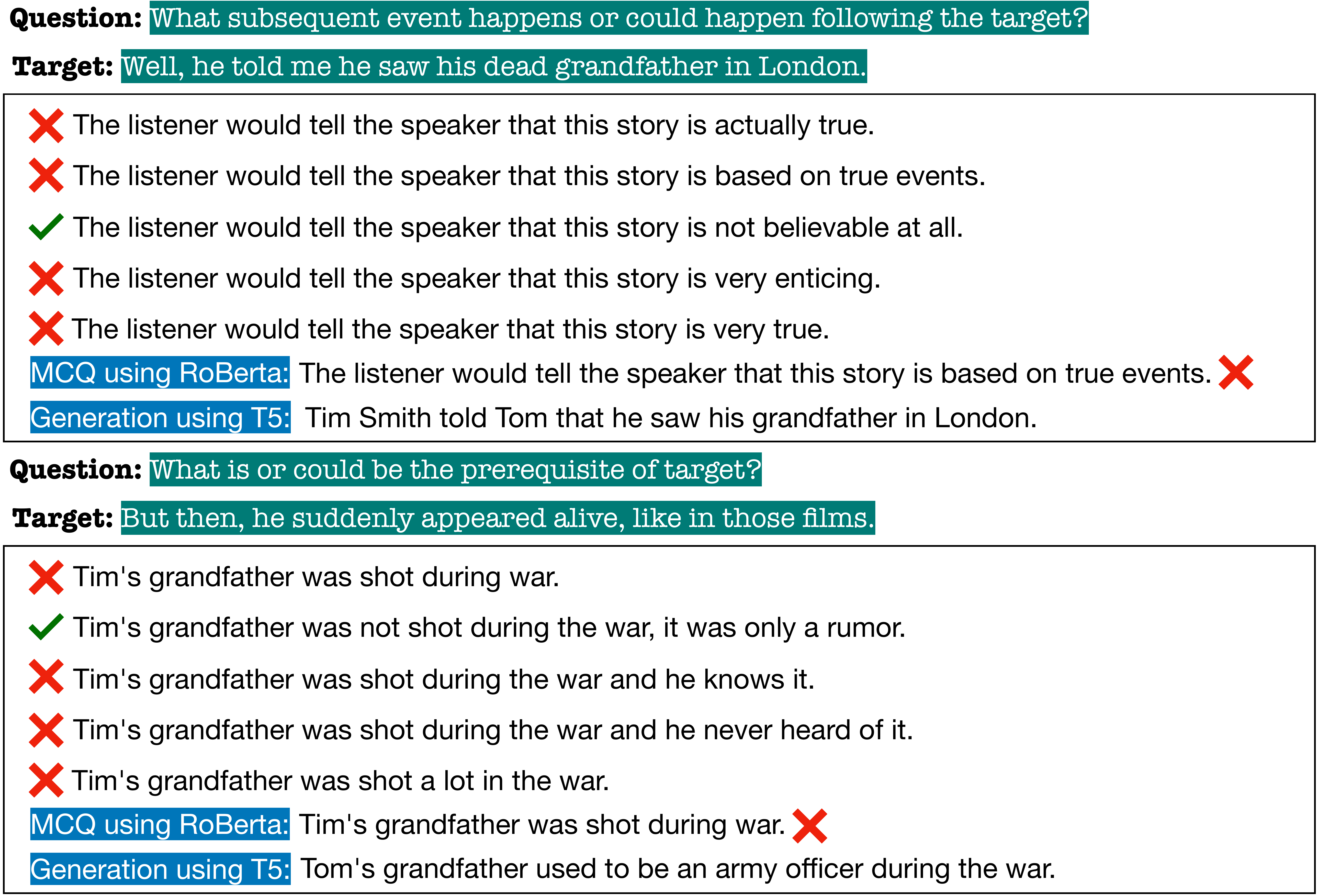}
        \label{fig:mcq12}
\end{subfigure}
\caption{Examples of some incorrect predictions by \code{RoBERTa} for the \dataset{}$_{MCQ}$ task.}
\label{fig:mcq11-12}
\end{figure}

\section{Additional Details on \dataset{}}
The total compensation for the complete annotation process of \dataset{} including all the manual labeling (\cref{sec:dataset}), and verification stages in AF (\cref{sec:altsc}) was USD $13,500$. The annotators were hired through a data annotation company. The total compensation was derived based on the country of residence of the annotators, as deemed by the company.

Being a dialogue-centric dataset, \dataset{} encompasses various aspects of human to human conversations such as temporal commonsense awareness in \cref{fig:mcq2}, \cref{fig:temporal-csk-sup}, physical commonsense in \cref{fig:mcq3}, general commonsense in \cref{fig:general_csk_sup}, and social commonsense in \cref{fig:social_csk_sup}.  In \cref{fig:mcq15}, commonsense is required to infer that a familiar face may look different to us if we meet that person after a long time. There could be other potential reasons why a person might look different to his/her friends such as facial surgery, sickness, makeup, etc. However, in this particular dialogue context, the most appropriate speculative cause of the target is meeting the person after a long time. Similarly in \cref{fig:mcq16}, the person hurries to the boarding gate as only 20 minutes is left before the flight takes off. Leveraging commonsense inference, we can infer that going to a place in a very short period requires us to rush.  In \cref{fig:mcq3}, \texttt{physical commonsense knowledge} is required to infer --- touching a hot element can burn our fingers and pans or microwaves are used for cooking.

\section{\dataset{}$_{NLG}$ Task: Extended Results}
\label{sec:appendix-ex}
We report BLEU1 scores~\cite{papineni2002bleu} in addition to the automatic evaluation metrics described in \Cref{sec:results}. We also report results for generative tasks with the \code{BART-large} ~\cite{lewis2019bart}, and \code{COMET}~\cite{Hwang2021COMETATOMIC2O} model. \code{COMET} is a commonsense generation model from free text input. It is a pre-trained \code{BART-large} model fine-tuned on the ATOMIC dataset~\cite{Hwang2021COMETATOMIC2O}. In our work, we have used all the models in two distinct ways -- i) with fine-tuning and ii) without fine-tuning on \dataset{}. The results are shown in \cref{tab:results-sup}, and \cref{tab:cqa-sup}. Surprisingly, despite being pre-trained on a large commonsense inference dataset, the fine-tuned \code{COMET} model fails to outperform both fine-tuned \code{T5} and \code{BART} in most of the experiments. This could be due to catastrophic forgetting triggered by disparate inputs, which are at odds with ATOMIC. Further research is needed to draw any conclusion.

The results of human evaluation of the models are illustrated in \cref{tab:human-eval-sup}. It can be seen that all the models perform almost similarly on \dataset{} and stand far from reaching human-level performance.

\paragraph{Fine-tuned vs non Fine-tuned Evaluations.}
All the models perform very poorly when they are not fine-tuned on \dataset{}.
The non fine-tuned models generate gibberish sentences across all five inference categories. The automatic and human evaluation results of these models are also reported in \cref{tab:results-sup} and \cref{tab:human-eval-sup}, respectively. 
The results confirm that fine-tuning is necessary for dialogue-level commonsense inference thus reaffirming the importance of our curated dataset \dataset{}. The non fine-tuned \code{COMET} produces very short outputs (1--3 words, akin to ATOMIC annotations) that are not readily comparable with \dataset{}, resulting in poor evaluation scores.

Finally, we provide some additional examples to depict the inference generation quality of the models in \cref{tab:examples-sup}.

\section{\dataset{}$_{MCQ}$: Extended Results, Quantitative and Qualitative Analysis}
\label{sec:appendix-alt}
For answer selection with generative models in \dataset{}$_{MCQ}$, we train \code{T5} and \code{Unified QA} models under three distinct settings: 1) \textbf{Setting 1:} train models only on instances with a single-answer, 2) \textbf{Setting 2:} train models only on instances with multiple-answers, 3) \textbf{Setting 3:} train models on the entire dataset comprising both single and multiple-answers. 

The performances of both the generative models \code{T5} and \code{Unified QA} on instances with multiple answers are very poor (see \cref{tab:appendix-alt}, \cref{tab:appendix-alt2} and \cref{fig:mcq13}, \cref{fig:mcq14}). Further, we can also see instances where the predicted answers by these models contradict (see \cref{fig:mcq14}). While \code{T5} surpasses \code{Unified QA} for Setting 3, \code{Unified QA} shines over \code{T5} for the other two settings. 

\paragraph{Performance of \code{ELECTRA} vs \code{RoBERTa}.}
We also extend upon the results reported earlier for \code{ELECTRA} and \code{RoBERTa} in \Cref{sec:results-mcq} for the single answer selection (Task 2.1) in \dataset{}$_{MCQ}$. The performance of \code{ELECTRA} is notably better than \code{RoBERTa} on this task. We reckon this could be due to the fact that we train our adversarial filtering (AF) method using \texttt{RoBERTa}. As such the efficacy of AF to prevent exposing stylistic artifacts to the discriminators is lesser for ELECTRA compared to RoBERTa. In other words, \code{ELECTRA} is more efficient than \code{RoBERTa} for the \dataset{}$_{MCQ}$ task due to its ability to better discriminate machine-generated negative answers from human-annotated true answers by leveraging stylistic artifacts as observed in \citet{zellers2018swag}.

Despite performing decently on the single answer selection task for \dataset{}$_{MCQ}$, \code{RoBERTa} does make mistakes in understanding some very interesting commonsense-based inferences such as the ones illustrated in \cref{fig:mcq11-12}. In these two examples, commonsense inference is required to detect the bluff by Tim Smith. Among other kinds of errors, we find \code{RoBERTa} failing to capture contextual commonsense cues such as in \cref{fig:mcq10} --- if a person wanting to buy new batteries is informed about the availability of batteries at photocopy stores, that person will search for photocopy stores instead of ad stores. 

\paragraph{Zero-shot Setting.} We also set up a zero-shot setting for Task 2.1 -- Single Answer Selection and Task 2.2 -- All Answers Selection. Under this setting, we only keep instances pertaining to cause, prerequisite, and emotional reaction in the train, validation data while instances with subsequent event, and motivation are kept in the test data. All the models underperform in the zero-shot setting, as can be seen in \cref{tab:appendix-alt2}. Like the all and single answer(s) prediction, \code{T5} and \code{Unified QA} perform similarly. On the other hand, \code{ELECTRA}'s zero-shot performance surpasses that of \code{RoBERTa}. Notably, performance of \code{T5} and \code{Unified QA} only drop around 1\% in this setting, as compared to 3\% drop observed for \code{RoBERTa} and \code{ELECTRA}. Hence, it is fair to conclude that for the \dataset{}$_{MCQ}$ task, \code{T5} and \code{Unified QA} are more robust to zero-shot scenarios than \code{RoBERTa} and \code{ELECTRA}. In the case of zero-shot single answer prediction, the best model is \code{Unified QA} which outperforms \code{RoBERTa} and \code{ELECTRA} by 11\% and 15\% respectively. 

\paragraph{Performance on Single- vs Multi-answer Instances.} It is evident from \cref{tab:appendix-alt,tab:appendix-alt2}, that in both regular and zero-shot settings, all the models exclusively trained on single- and multi-answer instances perform better on single- and multi-answer test instances, respectively, as compared to models trained on both types of instances. This is likely a side-effect of the data imbalance between the single- and multi-answer instances ($\sim$86/14\%) in the training set which causes the scarce multi-answer instances to have confounding effect on the training process, degrading the performance on both types of test instances.

\paragraph{Performance of \dataset{}$_{NLG}$ vs \dataset{}$_{MCQ}$.}
We present the qualitative analysis for generative (\dataset{}$_{NLG}$) and discriminative (\dataset{}$_{MCQ}$) experiments in \cref{fig:mcq5}, \cref{fig:mcq6}, \cref{fig:mcq8}, \cref{fig:mcq9}, \cref{fig:mcq10}, \cref{fig:mcq4}, and \cref{fig:mcq7}. Except for \cref{fig:mcq10}, \code{RoBERTa} provides the accurate answer on all instances. Contrary to this, the performance of \code{T5} is far from being sublime on those samples for the \dataset{}$_{NLG}$ task. This depicts that the commonsense-based generative task \dataset{}$_{NLG}$ poses more challenge than the commonsense-based discriminative task \dataset{}$_{MCQ}$. We surmise this could happen due to two potential reasons --- 
\begin{enumerate}
    \item Machine-generated negative answers may carry stylistic biases~\cite{zellers2018swag}, thus making the task of discriminators easier.
    \item We collate the negative answers by generating counterfactual and contradictory sentences from the annotated true inferences. As a result, the generated negative answers are lexically very similar to the annotated sentences resulting in less diversity in the dataset. 
\end{enumerate}

\begin{table}[h!]
\centering
\small
\resizebox{0.9\linewidth}{!}{
\begin{tabular}{l|r}
\toprule
\textbf{Dataset}  & \textbf{RoBERTa-Large} \\
\midrule
Swag & 89.92 \\
HellaSwag &  85.20 \\
$\alpha$-NLI & 83.91 \\
Cosmos QA & 82.25 \\
Physical IQA & 79.40 \\
Social IQA & 77.12 \\
\midrule
\dataset{} & 83.28 \\
\bottomrule
\end{tabular}
}
\caption{Results of baseline models in other CSK datasets. Note: This result on \dataset{} using \code{RoBERTa-large} is obtained for only instances with single answer. }
\label{tab:other-csk-sup}
\end{table}

\section{\dataset{} vs Other Commonsense Datasets}
The key differences that set \dataset{} apart from the rest of the commonsense datasets are following:

\begin{itemize}
    \item To the best of our knowledge, \dataset{} is the only publicly available dialogue-centric commonsense inference dataset.
    \item The speculative nature of the questions posed to the annotators enforces employment of rich commonsense knowledge in the inferences, thereby, making \dataset{} commonsense-rich and, thus, difficult inferences for models without relevant commonsense knowledge.
    \item While the performance of the strong baseline models on \dataset{} for \dataset{}$_{MCQ}$ task are comparable (see \cref{tab:other-csk-sup}) with the performance on other available commonsense-based question-answering datasets, unlike the others, around 14\% of the instances in \dataset{} contain multiple correct inferences/answers. These are more challenging to the baselines, as can be seen in \cref{tab:appendix-alt}.
    \item Dialogue-centric commonsense inference/answer generation task, i.e., \dataset{}$_{NLG}$ is novel and hard to solve. Strong baselines, such as, T5, BART, and their checkpoints pre-trained on large external commonsense datasets, such as, ATOMIC and GLUCOSE, perform poorly at this task.
\end{itemize}

\section{Hyperparameter Details}
All models for the \dataset{}$_{NLG}$ generative tasks were trained with the Adafactor optimizer~\cite{shazeer2018adafactor} with a learning rate of 5e-6. The models  \dataset{}$_{MCQ}$ alternative selection were trained with the AdamW~\cite{loshchilov2018decoupled} optimizer with a learning rate of 1e-5. We used a batch size of 4 for all our experiments.

\section{Computational Resources}
The T5 Large and \code{GLUCOSE-T5} Large have 770M parameters each. The \code{RoBERTa-Large} and \code{ELECTRA-Large} have 355M and 335M parameters, respectively. We also use a \code{BART-Large} and \code{COMET-Large} models for more extensive experiments (\Cref{sec:appendix-ex}). Both the models have 406M parameters. We use a single RTX 8000 GPU for our experiments. All models were trained for 5 epochs. Training and inference for the generative tasks i.e., \dataset{}$_{NLG}$ require between 1.5-6 hours in this GPU. Training and inference for the alternative selection task i.e., \dataset{}$_{MCQ}$ require a total of 15 hours. Training and inference times are 40\% less for zero-shot setting experiments. 

\end{document}